\documentclass{article}

\usepackage[preprint]{neurips_2025}
\makeatletter
\renewcommand{\@noticestring}{Accepted to the International Joint Conference on Artificial Intelligence (IJCAI 2026).}
\makeatother

\usepackage[utf8]{inputenc}
\usepackage[T1]{fontenc}
\usepackage{microtype}
\usepackage{url}
\usepackage[hidelinks]{hyperref}
\usepackage{nameref}
\usepackage{graphicx}
\usepackage{subcaption}
\usepackage{booktabs}
\usepackage{tabularx}
\usepackage{array}
\usepackage{multirow}
\usepackage{colortbl}
\usepackage{xcolor}
\usepackage{amsmath}
\usepackage{amssymb}
\usepackage{amsfonts}
\usepackage{amsthm}
\usepackage{nicefrac}
\usepackage{enumitem}
\usepackage{xspace}

\setlist[itemize]{leftmargin=*, itemsep=2pt, topsep=3pt}
\setlist[enumerate]{leftmargin=*, itemsep=2pt, topsep=3pt}
\newcommand{\datasetname}{\texttt{MELLA}\xspace}

\title{MELLA: Bridging Linguistic Capability and Cultural Groundedness for Low-Resource Language MLLMs}

\author{%
  \textbf{Yufei Gao}$^{1,2}$ \quad
  \textbf{Jiaying Fei}$^{1}$ \quad
  \textbf{Nuo Chen}$^{3,*}$ \quad
  \textbf{Ruirui Chen}$^{4}$ \\
  \textbf{Guohang Yan}$^{1,*}$ \quad
  \textbf{Yunshi Lan}$^{2,*}$ \quad
  \textbf{Botian Shi}$^{1,5}$\thanks{Corresponding authors.} \vspace{4pt} \\
  $^{1}$Shanghai Artificial Intelligence Laboratory \\
  $^{2}$East China Normal University \\
  $^{3}$The Chinese University of Hong Kong, Shenzhen \\
  $^{4}$Institute of High Performance Computing (IHPC), Agency for Science, \\
  Technology and Research (A*STAR), Singapore \\
  $^{5}$Shanghai Innovation Institute \\
  \texttt{yfgao.a@gmail.com}
}

\begin{document}
\maketitle

\begin{abstract}
Multimodal Large Language Models (MLLMs) perform strongly in high-resource languages, yet often produce fluent but culturally ``thin'' descriptions in low-resource settings. We argue that this failure is not merely a linguistic limitation: culture-specific visual knowledge depends on native visual--textual alignments that translation-centric pipelines rarely provide.
We present \datasetname{}, a multimodal dataset across eight low-resource languages, designed to support linguistic fluency and cultural groundedness. \datasetname{} uses a dual-source strategy that combines native web image--alt-text pairs for culture-grounded supervision with generated-and-translated image descriptions for linguistically rich supervision, explicitly separating two learning signals often conflated in multilingual multimodal data.
Through controlled diagnostic fine-tuning on multiple MLLM backbones, we show that \datasetname{} mitigates cultural hallucination by helping models recognize and articulate culturally specific entities overlooked by translation-based adaptation. Our findings highlight data alignment, rather than model modification alone, as a path toward culturally grounded multimodal understanding in low-resource languages. 
Our dataset can be found at \url{https://opendatalab.com/applyMultilingualCorpus}.
\end{abstract}

\section{Introduction}

\label{sec:intro}

\begin{figure}[t]
    \centering
    \includegraphics[width=0.99\linewidth]{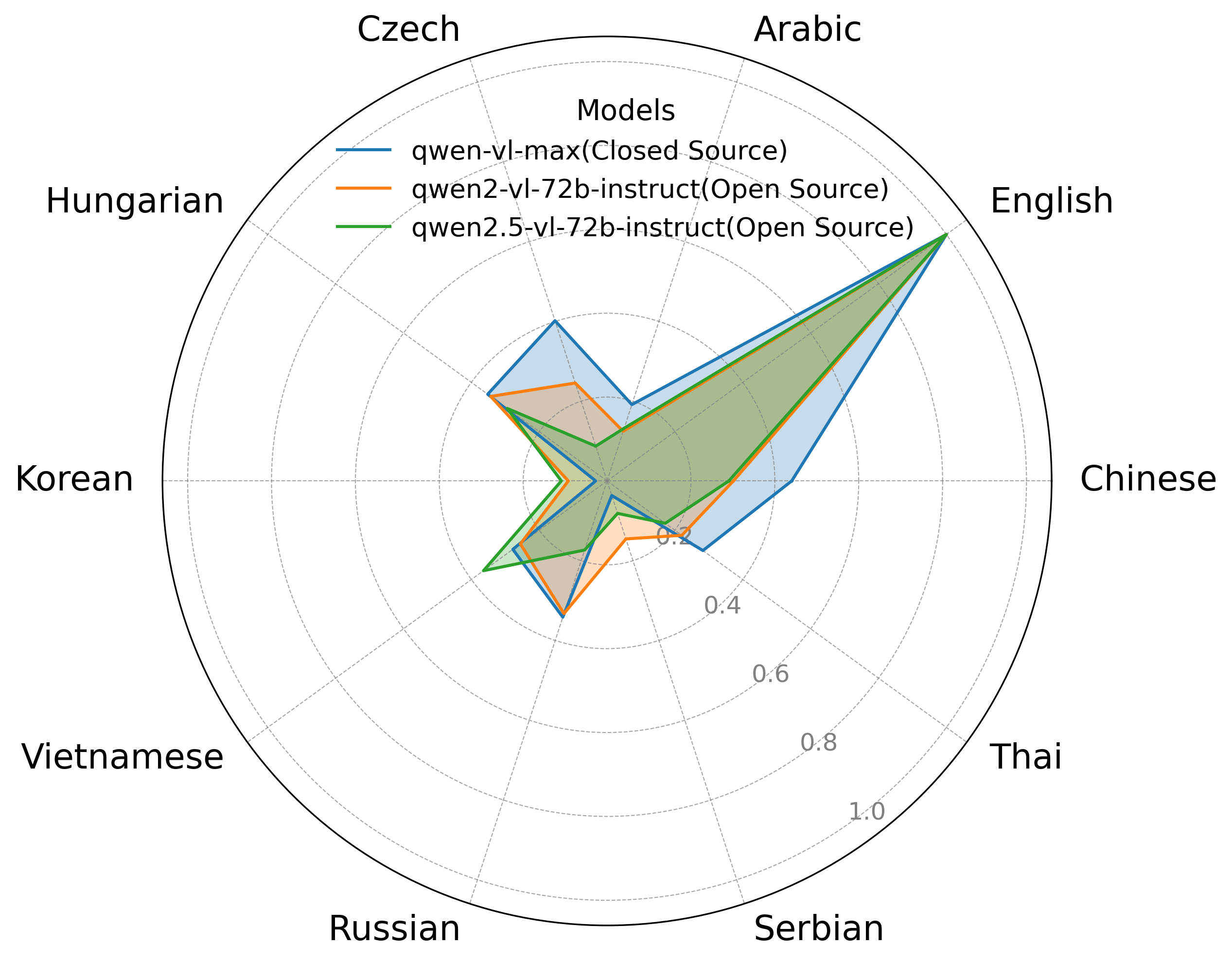}
    \caption{Image caption task performance on the COCO dataset~\citep{lin2015microsoftcococommonobjects} across multiple languages. Compared to GPT-4o~\citep{openai2024gpt4ocard}, most leading MLLMs achieve their highest BLEU~\citep{papineni-etal-2002-bleu} score in English.}
    \label{fig:compare}
\end{figure}

Multimodal Large Language Models (MLLMs) have achieved impressive performance in image understanding and generation~\citep{bai2025qwen25vltechnicalreport,chen2024expanding}, particularly in high-resource languages like
English, as illustrated in Figure~\ref{fig:compare}. However, in low-resource languages, these models suffer from a systematic failure mode: \textbf{the generation of fluent but culturally superficial descriptions that fail to recognize salient local entities}. For instance, a model may accurately describe ``a man in traditional clothing'' while failing to identify a significant public figure or ceremony that is immediately recognizable to native speakers. We refer to this phenomenon as \emph{cultural hallucination}: the production of culturally ``thin'' descriptions despite visually correct inputs.

\begin{figure}[ht]
    \centering
    \includegraphics[width=\linewidth]{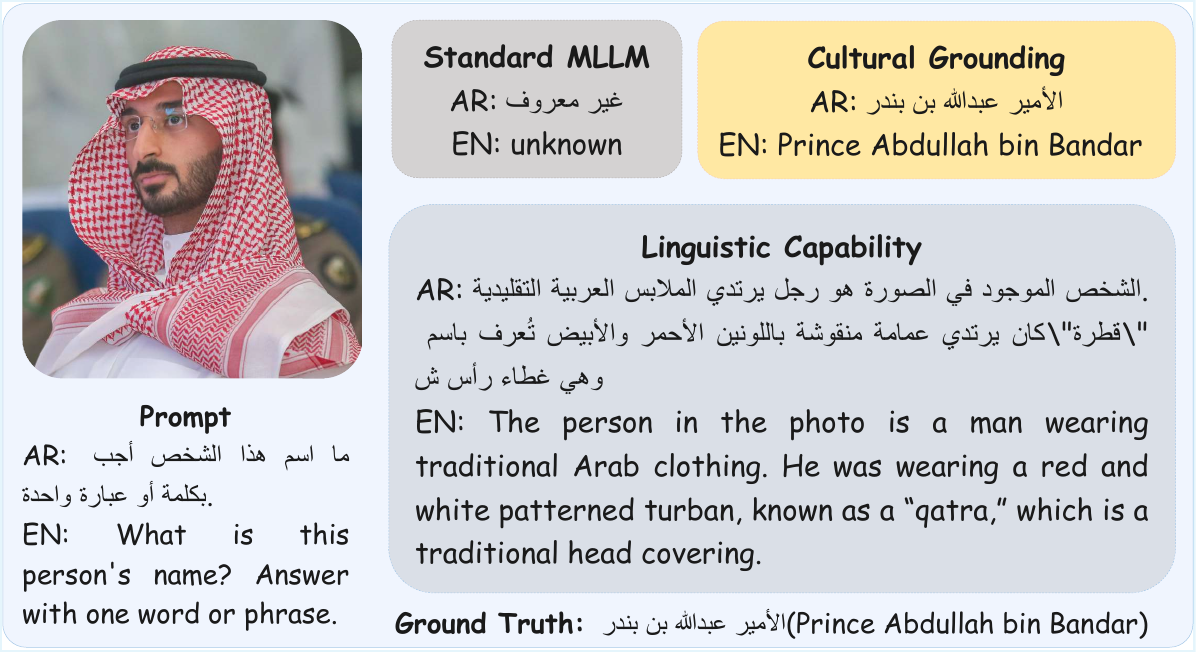}
    \caption{Standard MLLMs (e.g., InternVL2-8B, Qwen2-VL-7B) trained on generic datasets often fail to generate meaningful output due to limited visual-linguistic alignment. An MLLM with enhanced linguistic capability may produce detailed descriptions. However, only an MLLM enriched with cultural knowledge can accurately recognize the depicted celebrity. All conversations are expected to be in Arabic; ``EN'' provides translation for clarity.}
    \label{fig:linguistic-knowledge}
\end{figure}

% Cultural hallucination is not a linguistic fluency problem.
This failure cannot be explained by limited linguistic fluency alone. As shown in Figure~\ref{fig:linguistic-knowledge}, even models that produce fluent and detailed descriptions in low-resource languages may still overlook culture-specific visual knowledge. This indicates that \textbf{linguistic adaptation and cultural understanding are not equivalent}: a model can learn to speak a language fluently without learning how visual concepts are grounded in that culture. This leads to outputs that are plausible but culturally uninformative, undermining usability and trust for low-resource language users.

\begin{table}[ht]
\resizebox{\linewidth}{!}{%
\begin{tabular}{@{}lccll@{}}
\toprule
\rowcolor[HTML]{FFFFFF} 
\textbf{} & \multicolumn{1}{l}{\cellcolor[HTML]{FFFFFF}{\color[HTML]{1A1C1E} Multimodal}} & \multicolumn{1}{l}{\cellcolor[HTML]{FFFFFF}Cultural Awareness} & {\color[HTML]{1A1C1E} Linguistic Data Source} & {\color[HTML]{1A1C1E} Cultural Data Source} \\ \midrule
\rowcolor[HTML]{FFFFFF} 
{\color[HTML]{1A1C1E} SDRRL} & {\color[HTML]{1A1C1E} $\times$} & {\color[HTML]{1A1C1E} $\times$} & {\color[HTML]{1A1C1E} LLM-Gen+MT} & {\color[HTML]{1A1C1E} N/A} \\
\rowcolor[HTML]{FFFFFF} 
{\color[HTML]{1A1C1E} LexC-Gen} & {\color[HTML]{1A1C1E} $\times$} & {\color[HTML]{1A1C1E} $\times$} & {\color[HTML]{1A1C1E} Lexicon Translation} & {\color[HTML]{1A1C1E} N/A} \\
\rowcolor[HTML]{FFFFFF} 
{\color[HTML]{1A1C1E} Amharic LLaVA} & {\color[HTML]{1A1C1E} {\checkmark}} & {\color[HTML]{1A1C1E} $\times$} & {\color[HTML]{1A1C1E} MT Captions} & {\color[HTML]{1A1C1E} MT Captions} \\
\rowcolor[HTML]{EFEFEF} 
{\color[HTML]{1A1C1E} \textbf{Dual Source (Ours)}} & {\color[HTML]{1A1C1E} {\checkmark}} & {\color[HTML]{1A1C1E} {\checkmark}} & {\color[HTML]{1A1C1E} MLLM-Gen+MT} & {\color[HTML]{1A1C1E} Native Web Alt-text} \\ \bottomrule
\end{tabular}%
}

\caption{Comparison of multilingual enhancement approaches. Unlike methods that ignore image informativeness and rely on machine translation, our method promotes cultural awareness by sourcing data from \textbf{Native Web Alt-text}, authentic web image descriptions authored by individuals within specific cultural contexts.}
\label{tab:approach_comparison}

\end{table}

% Why translation-centric adaptation is insufficient.
A key commonality among existing multilingual MLLM adaptation methods is an implicit assumption: that cultural understanding can be induced indirectly through linguistic transfer. Methods such as SDRRL~\citep{zhang-etal-2024-enhancing-multilingual}, LexC-Gen~\citep{yong2024lexcgengeneratingdataextremely}, and Amharic LLaVA~\citep{andersland2024amharicllamallavamultimodal} extend language coverage primarily via translation-based or text-centric supervision, see Table~\ref{tab:approach_comparison}. While effective for improving surface-level fluency, these approaches \textbf{rarely introduce new native visual--textual alignments for culturally salient entities}. Consequently, they optimize linguistic expressiveness without addressing the underlying absence of culture-grounded multimodal data. 

% A data-alignment perspective on cultural grounding.
This reveals a fundamental limitation of translation-centric multilingual adaptation: in multimodal learning, cultural knowledge is \textbf{not a linguistic property} that can be transferred through language alone, \textbf{but a data alignment property} that depends on whether culturally salient entities ever appear as native visual--textual pairs during training. Consequently, this limitation is structural rather than algorithmic, and cannot be resolved by stronger translation models or increased model capacity.
This perspective aligns with classic semiotic accounts of visual meaning~\mbox{\citep{barthes1985rhetoric,geertz1973chapter}}. Images convey meaning not only through literal denotation but also through culturally coded connotation, a symbolic layer that translation alone cannot reliably recover. Without \textbf{native, culture-grounded supervision}, multilingual MLLMs are structurally biased toward producing \emph{``thin descriptions''} that omit culturally salient meaning.

To address this issue, we decompose image meaning into two components: a literal, objective denotation and a symbolic, culturally-coded connotation. Prior approaches to multilingual enhancement have primarily focused on the former. To bridge this gap, we explicitly introduce a dual objective for low-resource language MLLMs: (1) \textbf{Linguistic Capability}, which ensures fluency and nuanced expression, and (2) \textbf{Cultural Groundedness}, which enables understanding of culturally specific knowledge. Recognizing that the gap largely stems from an imbalance of culturally-relevant multimodal data across languages~\citep{romero2024cvqaculturallydiversemultilingualvisual}, we further propose a high-level, dual-source framework that integrates both a data collection strategy and a training objective to achieve this dual goal.

\begin{table*}[ht]
    \centering
    \scriptsize
    \renewcommand{\arraystretch}{1.5}
    \resizebox{0.98\textwidth}{!}{%
    \begin{tabular}{>{\raggedright\arraybackslash}p{1.0cm}|>{\raggedright\arraybackslash}p{3.4cm}|>{\raggedright\arraybackslash}p{1.7cm}|>{\raggedright\arraybackslash}p{0.9cm}|>{\raggedright\arraybackslash}p{8.6cm}}
        \hline
        \textbf{Dataset} & \textbf{Primary Goal} & \textbf{Low-Resource Focus} & \textbf{Cultural Focus} & \textbf{Data Curation Method} \\
        \hline
        WIT & Large-scale Pre-training & Incidental (100+ languages) & Incidental & Sourced from Wikipedia image-caption pairs across languages. \\
        \hline
        LAION-5B & Large-scale Pre-training \& Finetuning & Incidental (English-centric) & Incidental & Filtered Common Crawl based on CLIP score; alt-texts are unverified. \\
        \hline
        MTV-QA & Multilingual Text-centric VQA Benchmarking & Targeted & Incidental & Filtered Common Crawl based on OCR API; manually collect. \\
        \hline
        EXA-MS & Multilingual Exam Benchmarking & Targeted & Specific & Sourced from multilingual high school exam papers.  \\
        \hline
        CVQA & Cultural Benchmarking & Targeted & Specific & Local annotators manually collect images and create questions based on a guideline. \\
        \hline
        \textbf{MELLA (Ours)} & \textbf{Fine-tuning for Cultural \& Linguistic Skills} & \textbf{Targeted} & \textbf{Specific} & \textbf{Automated collection and annotation; Dual Source: 1) Native web alt-text for cultural Groundedness; 2) MLLM-generated descriptions for linguistic capability.} \\
        \hline
    \end{tabular}%
    }%
    \caption{Comparison of multimodal datasets: WIT~\citep{10.1145/3404835.3463257}, LAION-5B~\citep{schuhmann2022laion5b}, MTV-QA~\citep{tang2024mtvqa}, EXA-MS~\citep{das2024examsvmultidisciplinemultilingualmultimodal}, CVQA~\citep{romero2024cvqaculturallydiversemultilingualvisual}. }
    \label{tab:dataset_comparison}
    
\end{table*}

To instantiate the dual-source framework, we construct \datasetname \footnote{\url{https://opendatalab.com/applyMultilingualCorpus}}, the first initiative to address the dual challenges jointly. As Table \ref{tab:dataset_comparison} shows, \datasetname is unique in its motivation and data curation method. The construction and usage of \datasetname follow the proposed dual-source data strategy.  First, to instill \emph{cultural groundedness}, we curate native web corpora, extracting images along with their original HTML alt-text to form a knowledge-rich dataset $D_{know}$. This alt-text provides invaluable, human-authored context about culturally specific people, places, and objects.
Second, to foster \emph{linguistic capability}, we leverage a state-of-the-art MLLM to generate detailed English image descriptions, which are then translated into the target languages to create a linguistics-focused dataset $D_{ling}$. Experiments on two model backbones show clear improvements across both goals using our dataset, indicating the effectiveness of the dual-source framework.    

Our main contributions are:\begin{itemize}
    \item We propose a dual objective for low-resource language MLLMs, placing special emphasis on cultural awareness. To support this, we also introduce a dual-source strategy that offers high-level guidance toward fulfilling the dual objective. (Sec.~\ref{sec:methodology})
    \item We present \datasetname, a novel multimodal multilingual image-text dataset across eight low-resource languages as an instantiation of the dual-source strategy. (Sec.~\ref{sec:mella})
    % \item Extensive experiments across various model backbones demonstrate the effectiveness of our strategy, achieving significant improvements over existing methods. (Section~\ref{sec:exp})
    \item Through controlled diagnostic experiments, we show that translation-centric multilingual adaptation improves linguistic fluency but fails to recover culture-grounded visual knowledge, while our dual-source framework explicitly addresses this gap. (Sec.~\ref{sec:exp})

\end{itemize}

\section{Bridging Linguistic Capability and Cultural Groundedness}
\label{sec:methodology}

Motivated by the observation that low-resource multilingual MLLMs often produce fluent yet culturally ``thin'' descriptions, we formalize cultural hallucination as a failure to jointly acquire linguistic fluency and culture-grounded visual understanding. We then define a dual objective and propose a framework that explicitly bridges these two capabilities.

\subsection{Motivation}
\label{sec:decomposition}

Cultural hallucination arises when a model can describe what is visible in an image but fails to recognize what is culturally meaningful. To make this distinction precise, we draw from semiotics~\citep{barthes1985rhetoric,geertz1973chapter} and view image meaning as comprising two complementary components: a literal \emph{denotation} and a symbolic, culturally coded \emph{connotation}.
Formally, we decompose the total meaning $\mu$ of an image $I$ into:
\begin{equation}
    \mu(I) = (\mu_{den}(I), \mu_{con}(I)),
\end{equation}

% \noindent where $\mu_{\text{den}}$ corresponds to a ``thin description'' of what is explicitly visible, while $\mu_{\text{con}}$ captures the ``thick description'', the culturally embedded knowledge, identities, and social significance associated with the image.
\noindent where \emph{denotation} ($\mu_{\text{den}}$) corresponds to a ``thin description'' of what is explicitly visible, while \emph{connotation} ($\mu_{\text{con}}$) captures the ``thick description'', the culturally embedded knowledge, identities, and social significance associated with the image.

Prevailing multilingual MLLM adaptation methods (Table~\ref{tab:approach_comparison}), which primarily rely on translating English-centric image--text datasets, target $\mu_{\text{den}}$. As a result, such models may generate fluent descriptions of scenes while systematically failing to recover culturally salient entities (e.g., local public figures, ceremonies, or region-specific artifacts). This limitation reflects the core argument of this work: \textbf{without native visual--textual alignments, cultural knowledge cannot be recovered through linguistic transfer alone}.

\subsection{Dual Objective}
\label{sec:dual_objective}

To bridge the $\mu_{\text{den}}$--$\mu_{\text{con}}$ gap that translation-centric methods fail to address, we formalize two capabilities that an MLLM must acquire to work in low-resource settings.

\noindent \textbf{Objective 1: Linguistic Capability.}
We define linguistic capability as the model's ability to generate fluent and structurally accurate text in a target language. Formally, we denote this capability as:
\begin{equation}
    f_{\text{ling}}: (I, L) \rightarrow T_{\text{den}}^{L},
\end{equation}
where $L$ is the target language and $T_{\text{den}}$ is a textual realization of the denotative meaning $\mu_{\text{den}}(I)$. This capability corresponds to producing a ``thin description'' and requires mastery of vocabulary and grammar in language $L$.

\paragraph{Objective 2: Cultural Groundedness.}
We define cultural groundedness as the model's ability to recognize and articulate culturally specific knowledge embedded in an image. This capability is formalized as:
\begin{equation}
    f_{\text{cult}}: (I, L) \rightarrow T_{\text{con}}^{L},
\end{equation}
where $T_{\text{con}}$ represents a textual expression of the connotative meaning $\mu_{\text{con}}(I)$. Unlike linguistic capability, this function is difficult to learn from translation-based supervision alone, as it depends on exposure to authentic, culture-grounded visual--textual associations.

Together, these two objectives highlight that linguistic fluency and cultural grounding constitute \textbf{distinct but complementary learning signals}, both of which are necessary to mitigate cultural hallucination in multilingual MLLMs.

\subsection{Dual-source Framework}
\label{sec:framework}

To operationalize the dual objective, we propose a framework based on a dual-source data strategy that explicitly targets each capability.

\paragraph{Dual-source Data Strategy.}
Existing methods struggle to address $\mu_{\text{con}}$ primarily due to the scarcity of aligned, culturally relevant multimodal data for low-resource languages. To overcome this bottleneck, we construct a training corpus $D$ by combining two complementary data sources:
\begin{equation}
    D = D_{\text{ling}} \cup D_{\text{know}}.
\end{equation}

The first source, $D_{\text{ling}} = \{(I_i, T_{\text{den},i})\}_{i=1}^{M}$, is a linguistics-focused dataset designed to support $f_{\text{ling}}$. Each pair consists of an image and a fluent denotative description originally generated in English by a strong MLLM and subsequently translated into the target language $L$. This source provides rich syntactic and lexical supervision for low-resource language generation.

The second source, $D_{\text{know}} = \{(I_j, T_{\text{con},j})\}_{j=1}^{N}$, is a culture-grounded dataset designed to support $f_{\text{cult}}$. It is constructed from authentic in-culture sources (e.g., native web corpora), where images are paired with human-authored alt-text reflecting culturally specific knowledge and usage. Unlike translated captions, these descriptions encode native visual--textual alignments that are essential for learning $\mu_{\text{con}}$.

By jointly training on $D_{\text{ling}}$ and $D_{\text{know}}$, the model is encouraged to allocate representational capacity to both linguistic fluency and cultural grounding within a unified parameter space. This dual-source framework directly addresses the structural limitation identified in the introduction: \textbf{cultural grounding in multimodal models is fundamentally a data alignment problem rather than a purely linguistic one}.

\begin{figure*}[t]
    \centering
    \includegraphics[width=0.95\textwidth]{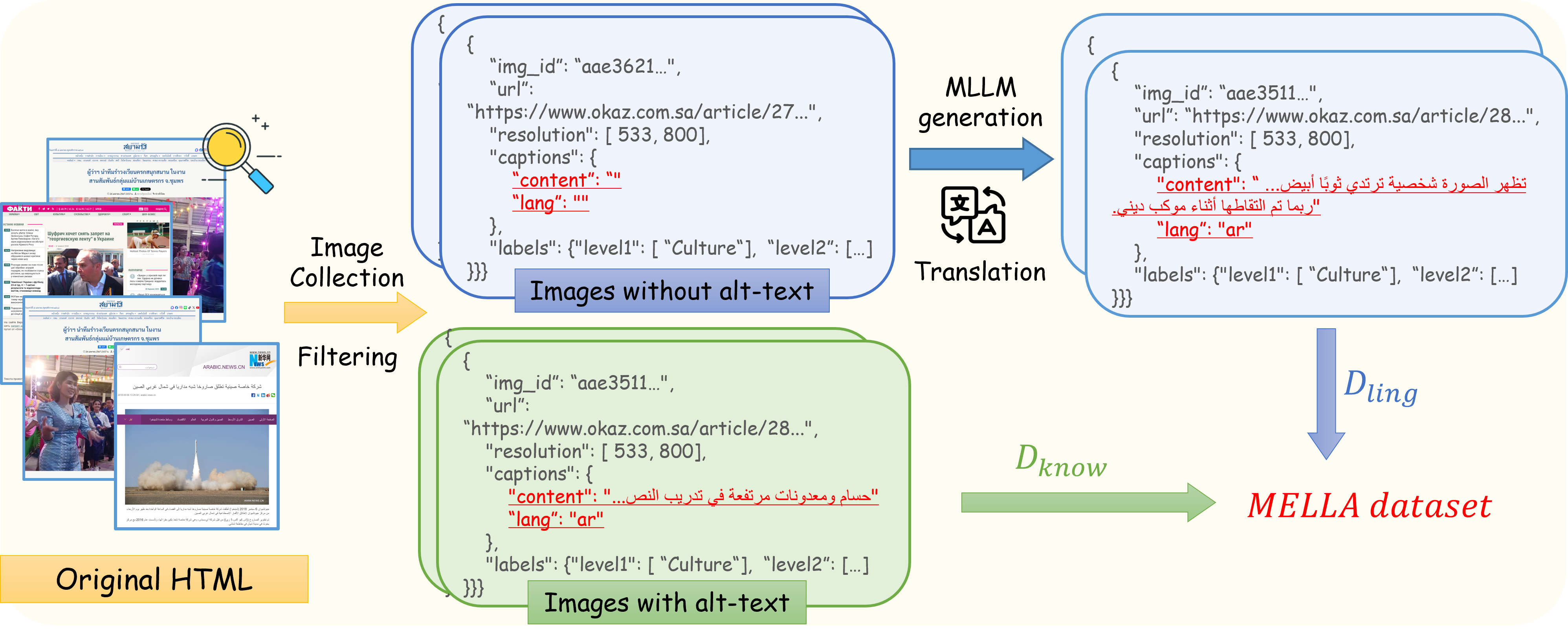}
    \caption{Data collection pipeline for \datasetname. Native web images with alt-text are used to construct the culture-grounded dataset ($D_{know}$). For images without native descriptions, we generate denotative captions in English using a high-resource MLLM and translate them into target languages to form the linguistic dataset ($D_{ling}$).}
    \label{fig:data_collection_pipeline}
\end{figure*}

\section{\datasetname: Instantiating the Framework}
\label{sec:mella}

\datasetname{} (\textbf{M}ultilingual \textbf{E}nhancement for \textbf{L}ow-resource \textbf{LA}nguage MLLMs) is a dual-source multimodal dataset that directly instantiates the framework introduced in Section~\ref{sec:framework}. This section describes the data construction process, quality assurance principles, and dataset statistics. We follow standard supervised fine-tuning (SFT) with cross-entropy loss for all experiments, as our primary contribution lies in data design rather than model optimization. 
% ; training objectives and prompt details are provided in Section~\ref{subsec:exp_setup} and Appendix~\ref{apdx: training details}

\subsection{Dataset Construction}
\label{sec:dataset}

The construction of \datasetname{} follows three stages: (1) image collection from native web sources, (2) acquisition of culture-grounded and denotative textual supervision, and (3) quality assurance and filtering (Figure~\ref{fig:data_collection_pipeline}).

\noindent \textbf{Image Collection}
We focus on eight low-resource languages: Arabic (AR), Czech (CS), Hungarian (HU), Korean (KO), Russian (RU), Serbian (SR), Thai (TH), and Vietnamese (VI), which are underrepresented in existing multimodal datasets~\citep{tang2024mtvqabenchmarkingmultilingualtextcentric,10.1145/3404835.3463257}. To obtain culturally situated visual content, we crawl $24$ high-traffic, language-native websites spanning multiple domains (e.g., news media, government portals, educational resources, forums, and commercial platforms). This multi-domain design mitigates domain bias and improves coverage of everyday cultural contexts.
% The complete website list is provided in Appendix~\ref{apdx: weblist}.

Images are extracted from HTML pages and automatically categorized into $4$ coarse and $22$ fine-grained semantic categories using InternVL-1.5-25.5B~\citep{chen2024internvl}. We apply resolution filtering, deduplication, and content-based heuristics to remove low-quality or redundant images. 
% Detailed filtering criteria are described in Appendix~\ref{apdx:image_filter}. 
The final corpus contains approximately $6.82$ million images.

\noindent \textbf{Alt-text for Cultural Groundedness ($D_{know}$)}
To support cultural groundedness ($f_{\text{cult}}$), we construct $D_{know}$ from native web alt-text. Authored by users within the target culture, alt-text often contains culturally situated entities (e.g., local figures, events, and region-specific artifacts), serving as a valuable albeit noisy and linguistically sparse source of aligned visual--textual supervision~\citep{sharma2018conceptual,Chintalapati_2022}. For each target language $L$, we extract alt-texts written in the same language as the hosting web page and pair them with their corresponding images:
\begin{equation}
D_{know}^{L} = \{(I_i^{L}, T_{\text{con},i}^{L})\}_{i=1}^{N}.
\end{equation}
Language consistency is verified using HTML metadata and fastText~\citep{joulin2016bag}, ensuring 100\% target language purity. We emphasize that $D_{know}$ is not assumed to be noise-free; instead, it provides native visual--textual alignments that are largely absent from translation-centric pipelines, directly operationalizing our claim that cultural grounding is a data alignment property.

\noindent \textbf{High-Resource Pivot for Linguistic Capability ($D_{ling}$)}
To support linguistic capability ($f_{ling}$), we construct $D_{ling}$ using a \textbf{High-Resource Pivot} strategy. Rather than generating captions directly in low-resource languages, often resulting in poor fluency or hallucination, we first generate detailed, denotative descriptions in English using a strong MLLM (e.g., GPT-4o~\citep{openai2024gpt4ocard} or Qwen2.5-VL~\citep{bai2025qwen25vltechnicalreport}), where visual reasoning is most reliable.

The generation prompts are explicitly constrained to describe only visually observable content, avoiding identity inference or culturally specific speculation.
% (prompt details in Appendix~\ref{apdx:image_desc}). 
The English captions are then translated into the target languages using high-quality machine translation systems (DeepL or Google Translate, depending on language support), followed by human spot-checking. Translation quality is evaluated using COMET-Kiwi~\citep{rei-etal-2022-cometkiwi}, achieving an average score of $0.75$. The resulting dataset is:
\begin{equation}
D_{ling}^{L} = \{(I_i^{L}, T_{\text{den},i}^{L})\}_{i=1}^{M}.
\end{equation}

\begin{table}[t]
    \centering
    {\small
    \setlength{\tabcolsep}{4pt}
    \renewcommand{\arraystretch}{1.08}
    \begin{tabular}{lccc}
    \toprule
    \multicolumn{1}{l}{\textbf{Statistic}} & \textbf{Number} & \textbf{Size (GB)} & \textbf{Avg. Length} \\ \midrule
    \multicolumn{1}{l}{Total of \(D\) } & 6816029 & 2153.924 & - \\ \midrule
    \multicolumn{1}{l}{\(D_{know}\)} & 2729891 & 1244.714 & 14 \\
    \(D_{know}-AR\) & 317954 & 77.33 & 22 \\
    \(D_{know}-CS\) & 364571 & 96.749 & 8 \\
    \(D_{know}-HU\) & 266889 & 203.147 & 11 \\
    \(D_{know}-KO\) & 367621 & 183.04 & 30 \\
    \(D_{know}-RU\) & 623140 & 148.44 & 9 \\
    \(D_{know}-SR\) & 271731 & 108.488 & 6 \\
    \(D_{know}-TH\) & 214627 & 169.52 & 8 \\
    \(D_{know}-VI\) & 303358 & 258.00 & 17 \\ \midrule
    \multicolumn{1}{l}{\(D_{ling}\)} & 4086138 & 909.21 & 258 \\
    \(D_{ling}-AR\) & 336321 & 79.771 & 256 \\
    \(D_{ling}-CS\) & 513795 & 92.013 & 260 \\
    \(D_{ling}-HU\) & 548428 & 133.221 & 263 \\
    \(D_{ling}-KO\) & 575581 & 135.951 & 251 \\
    \(D_{ling}-RU\) & 497414 & 109.382 & 251 \\
    \(D_{ling}-SR\) & 521856 & 88.811 & 261 \\
    \(D_{ling}-TH\) & 542863 & 119.142 & 261 \\
    \(D_{ling}-VI\) & 549880 & 150.919 & 261 \\ \bottomrule
    \end{tabular}
    }
    \caption{Statistics of the \datasetname dataset across data sources and languages.}
    \label{tab:dataset_statistics}
\end{table}

% \begin{figure}[t]
%     \centering
%     \includegraphics[width=\linewidth]{imgs/mella_picnew.png}
%     \caption{Category distribution of the \datasetname dataset.}
%     \label{fig:mella_overview}
% \end{figure}

\begin{table}[t]
\centering
% \scriptsize
\setlength{\tabcolsep}{5pt}
\renewcommand{\arraystretch}{1.05}
\begin{tabular}{@{}l r l r@{}}
\toprule
\multicolumn{2}{c}{\textbf{Culture} (31.04\%)} &
\multicolumn{2}{c}{\textbf{Scenes} (41.12\%)} \\
\cmidrule(lr){1-2}
\cmidrule(l){3-4}
\textbf{Subcategory} & \textbf{Ratio} &
\textbf{Subcategory} & \textbf{Ratio} \\
\midrule
Text        & 8.09\%  & Outdoor          & 11.49\% \\
Technology  & 7.64\%  & Indoor           & 9.04\%  \\
Folk Custom & 5.22\%  & Urban            & 7.77\%  \\
Sports      & 4.13\%  & Natural Scenery  & 5.52\%  \\
Religion    & 3.15\%  & Historical Sites & 3.71\%  \\
Art         & 2.82\%  & Rural            & 3.59\%  \\
\midrule
\multicolumn{2}{c}{\textbf{General} (13.32\%)} &
\multicolumn{2}{c}{\textbf{People} (14.52\%)} \\
\cmidrule(lr){1-2}
\cmidrule(l){3-4}
\textbf{Subcategory} & \textbf{Ratio} &
\textbf{Subcategory} & \textbf{Ratio} \\
\midrule
Food           & 4.33\% & Adult   & 9.01\% \\
Transportation & 2.93\% & Child   & 2.00\% \\
Industry       & 2.21\% & Elderly & 1.58\% \\
Animals        & 2.19\% & Youth   & 1.32\% \\
Plants         & 1.65\% & Infant  & 0.61\% \\
\bottomrule
\end{tabular}
\caption{Category distribution of the \datasetname dataset.}
\label{tab:mella_overview}
\end{table}

\noindent \textbf{Quality Assurance and Filtering}
We apply a rigorous two-stage quality assurance process. First, automated filters remove samples with language mismatch, encoding errors, duplicated content, or weak image--text relevance. Second, native speakers manually review random subsets of each language to identify residual linguistic or cultural errors. If the rejection rate exceeds $5$\%, the filtering rules are further refined and the cleaning pipeline is reapplied.
% Full protocols and statistics are reported in Appendix~\ref{apdx:qa}.

\subsection{Dataset Statistics and Release}

Table~\ref{tab:dataset_statistics} summarizes the dataset scale, and Table~\ref{tab:mella_overview} shows the category distribution. The dataset contains $6.8$M image--text pairs across eight languages, covering diverse topics and semantic categories. Following responsible data practices, we plan to release \datasetname{} under the \textbf{CC BY 4.0} license. The released data will include image URLs, native alt-text, generated denotative captions, and metadata (e.g., resolution and category labels).

\section{Experiments}
\label{sec:exp}

We conduct a data-centric, diagnostic evaluation to examine whether \datasetname mitigates the \emph{cultural hallucination} gap in low-resource multilingual MLLMs.
Rather than targeting state-of-the-art performance on public benchmarks, our goal is to isolate whether culturally grounded visual knowledge can be acquired when native visual--textual alignments are introduced.
Accordingly, our evaluation emphasizes controlled comparisons over leaderboard-style benchmarking, and complements large-scale multilingual models~\citep{dash2025ayavision,yue2025pangea} that prioritize breadth and scale.
% While large-scale multilingual models such as AyaVision and Pangea demonstrate strong cross-lingual generalization,
% their training objectives and data construction focus on breadth and scale rather than isolating culture-grounded visual knowledge.
% A direct leaderboard-style comparison would therefore confound data scale, task formulation, and supervision type.
% We instead include representative translation-centric baselines (e.g., SDRRL, mBLIP) that align with our controlled diagnostic setting.
Concretely, our experiments address three questions:
(1) Can \datasetname reduce culturally ``thin'' descriptions while preserving fluent generation in target languages?
(2) How do the two data sources ($D_{know}$ and $D_{ling}$) contribute complementary learning signals for cultural grounding and linguistic fluency?
(3) Under a controlled setting, how does joint dual-source training compare with staged or translation-centric adaptation pipelines?
Together, these analyses support the view that multilingual multimodal enhancement requires balancing two distinct learning signals: structural fluency and culture-grounded visual knowledge (cf. Section~\ref{sec:decomposition}).

% Concretely, we study four questions: (1) Does fine-tuning on \datasetname reduce culturally ``thin'' descriptions while maintaining fluent generation in the target languages? (2) What complementary roles do the two data sources ($D_{know}$ and $D_{ling}$) play? (3) How should the dual-source data be utilized during training to balance cultural grounding and linguistic fluency? (4) How does \datasetname compare against representative multilingual adaptation pipelines under the same controlled setting? By isolating linguistic versus cultural supervision, our experiments support the view that multimodal multilingual enhancement should be treated as a \emph{balanced acquisition of two distinct learning signals}: structural fluency and culture-grounded visual knowledge (cf. Section~\ref{sec:decomposition}).

% We conduct a data-centric, diagnostic evaluation to analyze whether \datasetname mitigates the \emph{cultural hallucination} gap in low-resource multilingual MLLMs.
% We emphasize that our primary goal is not to establish state-of-the-art performance on existing public benchmarks, 
% but to diagnose whether culturally grounded visual knowledge can be acquired when native visual--textual alignments 
% are introduced. Accordingly, our evaluation is designed to isolate data effects under a controlled task formulation, 
% rather than to measure cross-task generalization.

\begin{table*}[!ht]
\centering
\renewcommand{\arraystretch}{0.952} 
\setlength{\aboverulesep}{0.5pt}   
\setlength{\belowrulesep}{0.42pt}   
\resizebox{\linewidth}{!}{
\begin{tabular}{cccccccccc}
\toprule
\rowcolor[HTML]{F2F3F5}
\textbf{Backbone} & \textbf{Training} & \textbf{AR} & \textbf{SR} & \textbf{RU} & \textbf{CS} & \textbf{KO} & \textbf{TH} & \textbf{VI} & \textbf{HU} \\
\midrule
\rowcolor[HTML]{F0F4FF}
\multicolumn{10}{c}{\cellcolor[HTML]{F0F4FF}\textit{\textbf{Keyword Accuracy on $D_{know}$ (Cultural Groundedness)}}} \\
 & Base & 2.46 & 0.56 & 1.24 & 1.10 & 0.50 & 3.72 & 0.78 & 4.39 \\
InternVL2-8B             & SDRRL & 2.39 & 0.33 & 1.22 & 1.37 & 1.02 & 3.38 & 1.00 & 2.00 \\
             & \datasetname & 6.26$\pm$0.69 & 3.07$\pm$0.72 & 8.37$\pm$1.50 & 15.56$\pm$0.47 & 5.06$\pm$0.82 & 4.50$\pm$0.51 & 2.50$\pm$0.75 & 5.57$\pm$0.60 \\
\cmidrule(l){2-10}
 & Base & 1.56 & 0.80 & 3.12 & 2.89 & 2.00 & 4.55 & 0.32 & 2.16 \\
Qwen2-VL-7B            & SDRRL & 0.01 & 0.66 & 0.45 & 1.78 & 0.01 & 2.86 & 0.15 & 1.57 \\
            & \datasetname & 2.23$\pm$0.72 & 1.13$\pm$0.64 & 3.26$\pm$1.25 & 4.90$\pm$0.65 & 4.13$\pm$0.72 & 4.97$\pm$0.89 & 0.65$\pm$0.78 & 2.92$\pm$0.85 \\
\midrule
\rowcolor[HTML]{F0F4FF}
\multicolumn{10}{c}{\cellcolor[HTML]{F0F4FF}\textit{\textbf{BLEU on $D_{ling}$ (Linguistic Fluency Sanity Check)}}} \\
 & Base & 1.79 & 1.05 & 5.56 & 1.31 & 2.56 & 0.15 & 6.91 & 0.05 \\
InternVL2-8B             & SDRRL & 12.18 & 6.11 & 7.01 & 7.59 & 6.91 & 0.45 & 11.07 & 6.09 \\
             & \datasetname & 13.96$\pm$0.43 & 13.22$\pm$0.36 & 4.41$\pm$0.51 & 14.33$\pm$0.29 & 11.02$\pm$0.32 & 0.65$\pm$0.35 & 15.53$\pm$0.39 & 13.45$\pm$0.40 \\
\cmidrule(l){2-10}
 & Base & 2.45 & 0.60 & 3.24 & 2.37 & 1.48 & 0.32 & 8.17 & 3.40 \\
Qwen2-VL-7B            & SDRRL & 1.43 & 0.21 & 6.16 & 6.29 & 0.49 & 0.67 & 1.66 & 7.44 \\
            & \datasetname & 19.95$\pm$0.45 & 16.33$\pm$0.40 & 6.26$\pm$0.38 & 14.80$\pm$0.62 & 11.48$\pm$0.44 & 1.00$\pm$0.25 & 30.18$\pm$0.68 & 13.39$\pm$0.49 \\
\bottomrule
\end{tabular}
}
\caption{\textbf{Diagnostic evaluation on \datasetname held-out test sets.} Keyword Accuracy on $D_{know}$ measures culture-specific entity identification, while BLEU on $D_{ling}$ serves as a sanity check for linguistic fluency. Results highlight how different supervision signals affect complementary capabilities in multilingual MLLMs.
% ~\ref{apdx:full res}.
}
\label{tab: main results}
\end{table*}

\subsection{Experimental setup}
\label{subsec:exp_setup}

\paragraph{Training data.}
For each of the eight low-resource languages, we fine-tune models on a random subset of 80K--140K training pairs sampled from \datasetname.
% detailed statistics are reported in Appendix~\ref{apdx:training_data_statistics}. 
We intentionally avoid mixing \datasetname with external datasets to isolate the effect of our data design from confounding factors such as domain shift.

\paragraph{Diagnostic held-out evaluation.}
Because existing multilingual multimodal benchmarks do not simultaneously (i) cover our target low-resource languages and (ii) match our task formulation of producing detailed, culturally grounded descriptions, we construct diagnostic held-out test sets from \datasetname. Specifically, we sample 1{,}600 instances not used in training. For each language $L$, we include 100 examples from $D_{know}^{L}$ (culture-specific supervision) and 100 examples from $D_{ling}^{L}$ (linguistically rich supervision), totaling 200 test examples per language. We emphasize that this held-out evaluation is designed to \emph{diagnose cultural knowledge acquisition} within the same distribution as the training data, rather than to serve as a universal generalization benchmark. 
To further assess potential memorization effects, we additionally analyze train--test similarity and entity-frequency distributions. We observe low image-level nearest-neighbor overlap between training and test sets, and performance gains remain positive across entity-frequency buckets rather than concentrating only on high-frequency entities, suggesting that improvements are not solely explained by memorization of repeated samples or entity names.
% ~\ref{apdx:xm3600} showing that \datasetname fine-tuning does not degrade general captioning capability.

\paragraph{Evaluation metrics.}
Since $D_{know}$ and $D_{ling}$ probe different capabilities, we evaluate them using different metrics.
% Detailed metric definitions and prompts are provided in Appendix~\ref{apdx: evaluation details}.

\noindent \textbf{Cultural groundedness (on $D_{know}$).}
N-gram overlap metrics (e.g., BLEU/ROUGE) correlate poorly with cultural factuality: a model may generate a fluent caption while omitting or misidentifying the salient local entity. We employ \textbf{Keyword Accuracy} as a proxy for culture-specific entity recall, focusing on whether models recover salient entities grounded in the image. Following DeFactoNLP~\citep{reddy2018defactonlpfactverificationusing}, we extract TF--IDF keywords from the ground-truth alt-text to verify their presence in model outputs. We adopt TF--IDF rather than language-specific NER tools due to the limited availability and reliability of NER systems for low-resource languages~\citep{keraghel2024recentadvancesnamedentity}. 

\noindent \textbf{Linguistic fluency (on $D_{ling}$).}
To quantify whether models can produce fluent, structurally complete descriptions in the target low-resource languages, we report standard generation metrics on $D_{ling}$: \textbf{BLEU-4}~\citep{papineni-etal-2002-bleu}, \textbf{ROUGE-L}~\citep{lin-2004-rouge}, and \textbf{METEOR}~\citep{denkowski-lavie-2014-meteor}. We treat these metrics as \emph{sanity checks} for linguistic quality rather than as measures of cultural correctness. 

\paragraph{Baselines and comparable methods.}
We evaluate two widely used MLLM backbones, InternVL2-8B and Qwen2-VL-7B-Instruct, to ensure conclusions are not backbone-specific.
We compare: (i) \textbf{Base} (no fine-tuning), which reflects the default multilingual and cultural coverage of the backbone; and
(ii) \textbf{SDRRL}~\citep{zhang-etal-2024-enhancing-multilingual}, a representative translate-then-SFT multilingual adaptation pipeline that primarily targets linguistic adaptation via cross-lingual transfer and parallel resources, without explicit culture-grounded native supervision.
Rather than exhaustively comparing against all existing multilingual datasets or models, we focus on such translation-centric pipelines as representative baselines, enabling a controlled comparison aligned with our core hypothesis.
% We additionally report comparisons with mBLIP~\citep{geigle2024mblipefficientbootstrappingmultilingual} in Appendix~\ref{apdx: mblip}.

\paragraph{Implementation details.}
All experiments are implemented with DeepSpeed~\citep{10.1145/3394486.3406703} on two NVIDIA A100-SXM4-80GB GPUs.  Models are fine-tuned separately for each target language using the same training configuration and hyperparameter settings across all experiments.
% Full hyperparameters are in Appendix~\ref{sec:hyper}.

\subsection{Results}
\label{sec:results}

\subsubsection{Diagnostic results on the dual objective}
\label{sec: main results} Table~\ref{tab: main results} provides a diagnostic view of how different training strategies shape two distinct capabilities in multilingual MLLMs: culture-grounded entity recognition and fluent generation in low-resource languages.
 Rather than interpreting these numbers as a leaderboard comparison, we analyze the magnitude and direction of capability changes induced by different forms of supervision.

\noindent \textbf{Cultural grounding: recovering previously ignored entities.}
Across all languages, models trained without native cultural supervision exhibit \emph{near-zero keyword accuracy} on $D_{know}$, reflecting a systematic bias toward culturally ``thin'' descriptions. These models tend to describe images at a generic level (for example, ``a man in traditional clothing'') while failing to recognize culturally significant entities such as local public figures, ceremonies, or region-specific artifacts.
After fine-tuning on \datasetname, keyword accuracy increases substantially, typically by several-fold, indicating that native alt-text supervision enables models to \textbf{recognize and articulate} culturally grounded entities that were systematically ignored under translation-centric training.

\noindent \textbf{Linguistic fluency: preserving structurally complete descriptions.}
Importantly, introducing culture-grounded supervision does not come at the cost of linguistic fluency.
On $D_{ling}$, standard generation metrics such as BLEU and METEOR serve as \emph{sanity checks} for sentence completeness and surface-level fluency, rather than primary indicators of cultural understanding. 
Models fine-tuned on \datasetname{} maintain comparable or improved scores relative to their base counterparts across languages, suggesting that culture-grounded supervision does not degrade the model's ability to generate fluent and well-formed descriptions in low-resource languages.

\noindent \textbf{Limits of translation-centric adaptation.}
Translation-centric adaptation improves linguistic fluency but remains insufficient for cultural grounding.
SDRRL primarily targets cross-lingual \emph{linguistic} transfer and accordingly yields stronger fluency metrics on $D_{ling}$. However, on $D_{know}$ its Keyword Accuracy remains low and often close to that of base models, indicating that translation-based supervision alone does not reliably recover culture-specific entities grounded in the image. This contrast directly supports our central claim that \textbf{linguistic adaptation and cultural grounding are distinct, complementary learning signals}, and motivates incorporating native, culture-grounded multimodal supervision.

% SDRRL improves text-generation metrics on $D_{ling}$, consistent with its emphasis on cross-lingual linguistic transfer. However, its Keyword Accuracy on $D_{know}$ remains low and, in several languages, comparable to that of base models. Qualitative inspection reveals that SDRRL often substitutes culturally specific entities with generic descriptions or produces cross-lingual outputs, highlighting that translation-based supervision alone is insufficient to recover culture-specific visual knowledge. Together, these observations support our claim that \textbf{linguistic adaptation and cultural grounding constitute distinct and complementary learning signals}.

\begin{table}[t]
\centering
\resizebox{0.99\linewidth}{!}{%
\begin{tabular}{@{}cccccccccc@{}}
\toprule
\rowcolor[HTML]{F2F3F5} 
\textbf{Backbone} & \textbf{Method} & \textbf{AR} & \textbf{SR} & \textbf{RU} & \textbf{CS} & \textbf{KO} & \textbf{TH} & \textbf{VI} & \textbf{HU} \\ \midrule
% ================= KEYWORD ACCURACY =================
\multicolumn{10}{c}{\cellcolor[HTML]{F0F4FF}\textit{\textbf{Keyword Accuracy (on $D_{know}$)}}} \\ 
% InternVL2-8B 组
& $D_{ling}$ only & 3.20 & 0.56 & 2.80 & 1.80 & 0.72 & 5.10 & 1.10 & 3.50 \\
& $D_{know}$ only & 7.00 & 6.43 & 10.62 & 17.66 & 6.90 & 2.21 & 2.78 & 5.81 \\
& Two-stage & 7.01 & 5.46 & 13.48 & 21.09 & 8.00 & 2.29 & 3.56 & 6.32 \\
\multirow{-4}{*}{InternVL2-8B} & Joint (\datasetname) & 6.26 & 3.07 & 8.37 & 15.56 & 5.06 & 4.50 & 2.50 & 5.57 \\ \cmidrule(l){1-10} 
% Qwen2 组
& $D_{ling}$ only & 2.08 & 0.88 & 0.36 & 4.35 & 1.60 & 5.31 & 0.41 & 2.79 \\
& $D_{know}$ only & 1.26 & 1.86 & 3.09 & 2.67 & 4.63 & 1.84 & 1.46 & 2.29 \\
& Two-stage & 2.20 & 3.56 & 4.02 & 4.57 & 4.53 & 4.44 & 1.50 & 2.96 \\
\multirow{-4}{*}{Qwen2-VL-7B-Instruct} & Joint (\datasetname) & 2.23 & 1.13 & 3.26 & 4.90 & 4.13 & 4.97 & 0.65 & 2.92 \\ \midrule
% ================= METEOR =================
\multicolumn{10}{c}{\cellcolor[HTML]{F0F4FF}\textit{\textbf{METEOR (on $D_{ling}$)}}} \\
% InternVL2-8B 组
& $D_{ling}$ only & 37.90 & 17.29 & 14.81 & 15.59 & 29.39 & 35.10 & 33.41 & 16.16 \\
& $D_{know}$ only & 2.81 & 0.28 & 0.31 & 0.56 & 1.01 & 1.48 & 1.94 & 0.34 \\
& Two-stage & 13.65 & 0.31 & 0.27 & 0.52 & 1.38 & 1.76 & 1.81 & 0.37 \\
\multirow{-4}{*}{InternVL2-8B} & Joint (\datasetname) & 29.78 & 13.54 & 4.91 & 12.17 & 22.81 & 22.50 & 16.37 & 13.11 \\ \cmidrule(l){1-10} 
% Qwen2 组
& $D_{ling}$ only & 37.36 & 17.13 & 15.77 & 15.79 & 27.39 & 35.84 & 32.83 & 15.28 \\
& $D_{know}$ only & 2.13 & 0.04 & 0.40 & 0.49 & 0.81 & 1.02 & 1.50 & 0.22 \\
& Two-stage & 2.72 & 0.06 & 0.89 & 0.89 & 3.79 & 21.20 & 1.74 & 0.64 \\
\multirow{-4}{*}{Qwen2-VL-7B-Instruct} & Joint (\datasetname) & 36.89 & 13.88 & 5.36 & 12.88 & 23.74 & 34.63 & 28.66 & 12.72 \\
\bottomrule
\end{tabular}%
}
\caption{\textbf{Ablation: disentangling learning signals.} We compare fine-tuning with only $D_{ling}$ (linguistic supervision) or only $D_{know}$ (culture-grounded supervision), a two-stage pipeline ($ling\rightarrow know$), and joint training (\datasetname).}
\label{tab:ablation1}

\end{table}

\subsubsection{Ablation study}
Table~\ref{tab:ablation1} validates the complementarity of our dual-source design. Training on $D_{ling}$ alone yields strong METEOR but weak Keyword Accuracy, indicating fluent yet culturally ``thin'' descriptions. Conversely, $D_{know}$ alone improves keyword recall but severely degrades linguistic fluency, reflecting the stylistic sparsity of native alt-text. A two-stage approach partially recovers cultural recall but often harms linguistic metrics due to catastrophic forgetting: sequential LoRA training biases the model toward the later stage and makes it difficult to maintain a unified representation that supports both objectives. Joint training compels the model to allocate capacity for both learning signals within a shared parameter space, resulting in a more balanced outcome.
 % (see Appendix~\ref{apdx: training obj})

\subsubsection{Human evaluation}
Following the qualitative evaluation protocol in ShareGPT4V~\citep{chen2023sharegpt4vimprovinglargemultimodal}, we conduct human assessment to validate whether \datasetname reduces culturally ``thin'' descriptions in practice. We evaluate 100 samples comparing InternVL2-8B (Base) against InternVL2-8B fine-tuned on \datasetname, with 8 volunteers. The overall human evaluation results, reported in Table~\ref{tab:human-eval}, show a strong alignment with our quantitative findings.

\begin{table}[t]
\centering
\small
\setlength{\tabcolsep}{2.5pt}
\renewcommand{\arraystretch}{1.05}
\begin{tabular}{lcccccccc}
\toprule
\textbf{} & \textbf{AR} & \textbf{CS} & \textbf{HU} & \textbf{KO} & \textbf{RU} & \textbf{SR} & \textbf{TH} & \textbf{VI} \\
\midrule
MELLA Win (\%)    & 59.0 & 80.8 & 82.0 & 90.5 & 15.6 & 45.5 & 95.4 & 67.5 \\
Baseline Win (\%) & 22.3 & 0.0   & 3.9  & 1.3  & 69.3 & 34.1 & 0.0   & 17.2 \\
Comparable (\%)   & 18.7 & 19.2 & 14.1 & 8.2  & 15.1 & 20.4 & 4.6  & 15.3 \\
\bottomrule
\end{tabular}
\caption{Human evaluation results over 100 validation samples and 8 volunteers.}
\label{tab:human-eval}
\end{table}
\subsection{Further Analysis}
\label{sec:discussion}

\noindent \textbf{Low-resource fine-tuning does not harm high-resource performance.}
A common concern is that adapting MLLMs to low-resource languages may degrade performance in high-resource languages due to negative transfer or catastrophic forgetting. As shown in Table~\ref{tab:cross_lingual}, fine-tuning on individual low-resource languages consistently preserves English (EN) and Chinese (ZH) performance, with occasional small gains. This suggests that learning culture-grounded multimodal representations in low-resource settings does not interfere with, and may even support, general visual--language capabilities.

\begin{table}[h]
\centering
\setlength{\aboverulesep}{0.38pt}   
\setlength{\belowrulesep}{0.38pt}   
\small
\resizebox{0.96\linewidth}{!}{
\begin{tabular}{lcccc}
\toprule
\textbf{Training Language} & \textbf{EN Meteor} & \textbf{EN BLEU-4} & \textbf{ZH Meteor} & \textbf{ZH BLEU-4} \\
\midrule
Baseline (No FT) & 17.37 & 2.01 & 16.23 & 0.00 \\
\midrule
FT on Arabic (AR) & 17.28 & 2.25 & 17.21 & 0.06 \\
FT on Thai (TH) & 17.92 & 2.78 & 17.31 & 0.02 \\
FT on Korean (KO) & 17.59 & 2.54 & 16.80 & 0.02 \\
FT on Hungarian (HU) & 17.21 & 2.20 & 17.64 & 0.10 \\
\bottomrule
\end{tabular}
}
\caption{Cross-lingual transfer. Fine-tuning on one low-resource language maintains EN/ZH, indicating limited negative transfer.}
\label{tab:cross_lingual}

\end{table}

\noindent \textbf{Cultural hallucination manifests as fluent but entity-deficient descriptions.}
Error analysis reveals a consistent failure pattern we term \emph{cultural hallucination}: models generate fluent and visually plausible descriptions while omitting culturally salient entities that are obvious to native speakers. These errors are not random, but fall into three recurring categories: (i) \textbf{entity omission}, where specific local identities are ignored; (ii) \textbf{entity substitution}, where a culturally specific entity is replaced by a generic or foreign counterpart; and (iii) \textbf{over-generalization}, where culturally meaningful attributes are flattened into broad categories. These patterns explain why translation-centric adaptation yields linguistically fluent yet culturally ``thin'' outputs.

\noindent \textbf{Joint supervision is necessary to balance linguistic fluency and cultural grounding.}
The ablation results in Table~\ref{tab:ablation1} show that training on $D_{ling}$ alone improves linguistic fluency but fails to recover cultural entities, while training on $D_{know}$ alone increases entity recall at the expense of grammaticality. A staged training strategy ($ling\rightarrow know$) partially recovers cultural information but often harms linguistic metrics due to catastrophic forgetting under sequential optimization. In contrast, joint training consistently yields more balanced representations, supporting both fluent generation and culture-grounded entity recognition. This finding supports our design choice to treat linguistic and cultural supervision as a \emph{coupled} learning signal rather than independent stages.

\section{Conclusion}

This work addresses a persistent gap in low-resource multilingual MLLMs between linguistic fluency and cultural groundedness. We argue that this gap stems primarily from missing native, culture-grounded visual--textual alignments, rather than from model limitations. To this end, we propose a dual objective and a dual-source data strategy, instantiated as \datasetname{}, which explicitly separates linguistic supervision from culture-grounded knowledge. Through diagnostic experiments, we show that while translation-centric adaptation improves fluency, it remains insufficient for cultural grounding, whereas \datasetname{} enables models to recover culturally salient visual entities.

\newpage

\section*{Ethical Statement}

The dataset collected for this study is openly available under the Creative Commons Attribution 4.0 International License (CC BY 4.0). All collected images have undergone legal and safety review to ensure compliance with ethical and regulatory standards.

Our dataset is constructed from publicly available web content, and we plan to release the processed annotations and metadata under the CC BY 4.0 license. All images underwent rigorous filtering using an Image Moderation System to support privacy safeguards and remove harmful content, including violence, hate speech, and inappropriate material.

Web-crawled data may contain and perpetuate existing societal biases and stereotypes. Our cultural knowledge reflects web content, which may overrepresent dominant voices and not capture all perspectives within linguistic communities. We applied content filtering, but semantic biases may persist.

\section*{Acknowledgements}
The research was supported by Shanghai Artificial Intelligence Laboratory. This work was partially supported by the Natural Science Foundation of China (Project No. U23A20298).

\bibliographystyle{unsrtnat}
\bibliography{references}

\clearpage
\appendix
\section*{Appendix Index}
\textit{Note: Section numbers and titles are hyperlinked for easy navigation to the corresponding content.}
\begin{itemize} 
\item Data and Ethical Considerations
\begin{itemize}
    \item \hyperlink{sec:DataAvailability}{Data Availability, Release, and Governance Statement}
    \item \hyperlink{sec:Limitations}{Limitations}
    \item \hyperlink{sec:Risks}{Potential Risks (and Ethical Considerations)}
\end{itemize}

\item Section \ref{Related Work}: Related Work
\begin{itemize}
    \item Review of MLLMs, Multilingual Datasets, and Cultural Awareness.
\end{itemize}

\item Section \ref{apdx:data_collection_details}: \nameref{apdx:data_collection_details}
\begin{itemize}
    \item Appendix \ref{apdx: weblist}: Image Resource Website List.
    \item Appendix \ref{apdx:image_filter}: Image Filtering.
    \item Appendix \ref{apdx:image_desc}: Image Description Prompts.
    \item Appendix \ref{apdx:qa}: Quality Assurance and Filtering.
\end{itemize}

\item Section \ref{apdx: evaluation details}: \nameref{apdx: evaluation details}
\begin{itemize}
    \item Appendix \ref{apdx: Evaluation Prompts}: Evaluation Prompts.
    \item Appendix \ref{apdx: Automatic Metrics}: Automatic Metrics.
    \item Appendix \ref{apdx:tokenization}: Tokenization and Language Processing.
    \item Appendix \ref{apdx: Limitations of Keyword Accuracy}: Limitations of Keyword Accuracy.
\end{itemize}

\item Section \ref{apdx: training details}: \nameref{apdx: training details}
\begin{itemize}
    \item Appendix \ref{apdx: training obj}: Training Objective.
    \item Appendix \ref{sec:hyper}: Hyperparameters.
    \item Appendix \ref{apdx:prompt_pool}: Prompt Pool.
    \item Appendix \ref{apdx:training_data_statistics}: Training Data Statistics.
\end{itemize}

\item Section \ref{apdx:xm3600}: \nameref{apdx:xm3600}

\item Section \ref{Experiment results}: \nameref{Experiment results}
\begin{itemize}
    \item Appendix \ref{apdx:full res}: Full Main Results.
    \item Appendix \ref{apdx: mblip}: Comparison with mBLIP.
    \item Appendix \ref{apdx: Human evaluation results}: Human Evaluation Results.
\end{itemize}

\item Section \ref{apdx:perf_variation}: \nameref{apdx:perf_variation}

\item Section \ref{apdx:qual_human}: \nameref{apdx:qual_human}
\begin{itemize}
    \item Appendix \ref{apdx: human}: Human Evaluation Protocol.
    \item Appendix \ref{apdx:case_study}: Case Studies.
\end{itemize}

\end{itemize}

\newpage

\hypertarget{sec:DataAvailability}{}
\section*{Data Availability, Release, and Governance Statement} 
\label{sec: Data Availability Statement}
The dataset collected for this study is released under the Creative Commons Attribution 4.0 International License (CC BY 4.0). The release includes: (i) URLs of source web pages and image identifiers; (ii) two types of text annotations, namely native alt-text for $D_{know}$ and generated descriptions for $D_{ling}$; and (iii) metadata such as resolution and category tags. We do not redistribute raw copyrighted images; instead, we provide links, processed annotations, and metadata to support research use.

\datasetname is constructed from publicly accessible web content. All collected images have undergone legal and safety review to ensure compliance with ethical and regulatory standards. We also provide a takedown mechanism for content removal requests. Upon receiving a valid request, the corresponding samples will be promptly removed from future releases.

\hypertarget{sec:Limitations}{}
\section*{Limitations}
\label{sec: Limitations}
While our work focuses on eight low-resource languages, 
this represents only a small fraction of the world's low-resource languages. The selection was constrained by data availability and computational resources.

As shown in our ablation study~\ref{tab:ablation1}, two-stage training suffers from catastrophic forgetting. While our unified approach mitigates this, optimal integration strategies warrant further exploration, such as adding general datasets and developing new training methods.

Moreover, Although MELLA contains 6.8M image--text pairs in total, our experiments intentionally use 80K--140K samples per language to reflect realistic low-resource adaptation scenarios. 
We show that even at this scale, introducing native visual--textual alignments yields substantial gains in cultural grounding, suggesting that data quality and alignment can outweigh raw scale in low-resource settings.

\hypertarget{sec:Risks}{}
\section*{Potential Risks (and Ethical Considerations)}
\label{sec: Potential Risks (and Ethical Considerations)}
\textbf{Data Collection and Privacy.} All images underwent rigorous filtering using an Image Moderation System to support privacy safeguards and remove harmful content, including violence, hate speech, and inappropriate material. 

\textbf{Bias and Cultural Representation.} Web-crawled data may contain and perpetuate existing societal biases and stereotypes. Our cultural knowledge reflects web content, which may overrepresent dominant voices and not capture all perspectives within linguistic communities. We applied content filtering, but semantic biases may persist.

\section{Related Work}
\label{Related Work}
\subsection*{Multimodal Large Language Models}
Recent multimodal large language models (MLLMs) have demonstrated strong visual--language understanding, particularly in high-resource languages~\citep{team2023gemini, bai2025qwen25vltechnicalreport}. However, most open-source MLLMs remain limited in their multilingual coverage and robustness in low-resource settings. For example, Qwen-VL~\citep{bai2023qwenvlversatilevisionlanguagemodel} and LLaVA~\citep{liu2023visualinstructiontuning} lack dedicated mechanisms for low-resource language adaptation, while InternVL series~\citep{chen2024internvl, zhu2025internvl3exploringadvancedtraining} primarily supports English and Chinese. Multilingual instruction tuning, as adopted in LLaVA-1.5~\citep{liu2024llavanext, liu2023improvedllava}, improves instruction-following behavior but does not introduce new language-specific visual grounding. As a result, these models often produce fluent yet culturally generic outputs when deployed in low-resource languages. Our work does not modify model architectures; instead, we focus on addressing this limitation through data design.

\subsection*{Multilingual Multimodal Datasets}
Large-scale multimodal datasets are central to training and adapting MLLMs. Early datasets such as MSCOCO~\citep{lin2015microsoftcococommonobjects}, COCO-CN~\citep{li2019cococncrosslingualimagetagging}, and WIT~\citep{10.1145/3404835.3463257} primarily target high-resource languages, with limited culturally grounded coverage for low-resource languages. Recent efforts, including MTV-QA~\citep{tang2024mtvqabenchmarkingmultilingualtextcentric}, EXAMS-V~\citep{das2024examsvmultidisciplinemultilingualmultimodal}, and PM4Bench, expand multilingual evaluation but are largely text-centric or task-specific (e.g., OCR or exam-style reasoning), making them ill-suited for studying culturally grounded visual description.

Recent large-scale initiatives such as AyaVision~\citep{dash2025ayavision} and Pangea~\citep{yue2025pangea} primarily expand multilingual coverage through large-scale synthetic generation and translation. While effective for improving linguistic breadth, these datasets largely inherit the same limitation: culturally salient entities that never appear as native visual--textual pairs remain unlearnable. In contrast, \datasetname is explicitly designed to study a different failure mode: the absence of culture-grounded visual knowledge when native visual--textual alignments are missing.
As such, our goal is not to match their scale or breadth, but to isolate the effect of culturally native supervision under a controlled diagnostic setting.
% In contrast, \datasetname{} is explicitly designed to introduce such missing alignments through native web alt-text, enabling the study and mitigation of culturally ``thin'' descriptions in low-resource languages.

\subsection*{Cross-Lingual Transfer and Translation-Based Adaptation}
Cross-lingual transfer has been widely studied as a strategy for extending model capabilities to low-resource languages~\citep{andersland2024amharicllamallavamultimodal, yong2024lexcgengeneratingdataextremely, zhang-etal-2024-enhancing-multilingual}. Methods such as LexC-Gen~\citep{yong2024lexcgengeneratingdataextremely} and SDRRL~\citep{zhang-etal-2024-enhancing-multilingual} rely on translation or synthetic text generation to compensate for data scarcity, implicitly assuming that cultural understanding can be induced through linguistic transfer. While effective for improving surface-level fluency, these approaches rarely introduce new visual--textual alignments for culturally specific entities, limiting their ability to support culture-grounded understanding\citep{ding2026cross}. Our findings empirically validate this limitation and demonstrate that linguistic adaptation alone is insufficient for recovering culturally salient visual knowledge.

\subsection*{Cultural Awareness in Multimodal Models}
The cultural awareness of MLLMs has only recently attracted attention~\citep{pawar2025survey}. Benchmarks such as CVQA~\citep{romero2024cvqaculturallydiversemultilingualvisual} evaluate cultural knowledge through multiple-choice question answering, while CultureVLM~\citep{liu2025culturevlmcharacterizingimprovingcultural} focuses on improving cultural reasoning primarily in English. These efforts highlight the importance of cultural understanding but do not address how such knowledge should be acquired in low-resource languages. In contrast, our work identifies cultural grounding as a data alignment problem and provides a scalable dataset-driven solution that enables models to learn culture-specific visual concepts directly from native supervision.

% =========================
% Appendix: Data and Experiment Details
% =========================

\section{Data Collection Details}
\label{apdx:data_collection_details}

\subsection{Image Resource Website List}
\label{apdx: weblist}

\begin{table*}[ht]
\centering
\resizebox{\textwidth}{!}{%
\begin{tabular}{@{}cccc@{}}
\toprule
\textbf{ar} & \textbf{ru} & \textbf{ko} & \textbf{vi} \\
\midrule
\url{https://alghad.com} & \url{https://russian.rt.com} & \url{https://www.clien.net/service/} & \url{https://vietnambiz.vn/} \\
\url{https://albiladdaily.com} & \url{https://fakty.ua} & \url{https://www.healthfocus.co.kr/} & \url{https://daidoanket.vn/} \\
\url{https://www.okaz.com} & \url{https://old.day.kyiv.ua/uk} & \url{https://news.kbs.co.kr} & \url{https://www.qdnd.vn/} \\
\midrule
\textbf{th} & \textbf{hu} & \textbf{sr} & \textbf{cs} \\
\midrule
\url{https://siamrath.co.th/} & \url{https://www.blikk.hu/} & \url{https://www.kurir.rs/} & \url{https://www.patro.cz/} \\
\url{https://www.khaosod.co.th/} & \url{https://hvg.hu/} & \url{https://www.danas.rs/} & \url{https://cs.wikipedia.org/} \\
\url{https://buriram.mol.go.th/} & \url{https://prohardver.hu/} & \url{https://www.blic.rs/} & \url{https://www.euro.cz/} \\
\bottomrule
\end{tabular}%
}
\caption{Websites crawled for each target language.}
\label{tab:websites}
\end{table*}

Table~\ref{tab:websites} lists the websites we crawled from.

\subsection{Image Filtering}
\label{apdx:image_filter}
We observed that raw web images include low-resolution content, near-duplicates, irrelevant contexts, and potentially harmful or sensitive material. We therefore applied the following filtering steps.

\begin{itemize}
    \item \textbf{Resolution.} We retain only images whose width and height are both greater than 256 pixels.
    \item \textbf{Deduplication.} We apply a hierarchical deduplication strategy to reduce redundancy. We first remove exact duplicates with identical pixel-level content. We then apply pHash~\citep{zauner2010implementation} and remove near-duplicates by thresholding Hamming distance. Finally, we use a CNN feature extractor~\citep{krizhevsky2012imagenet} to remove semantically near-identical images at a finer granularity.
    \item \textbf{Safety and ethics.} We filter images containing sensitive or inappropriate content (e.g., violence, hate symbols, explicit content, or ads) using an Image Moderation System (IMS) API~\citep{tencent_ims}. We crawl only publicly accessible pages and respect website terms of service. We release only links, metadata, and processed annotations, as described in the \hyperlink{sec:DataAvailability}{Data Availability, Release, and Governance Statement}.
\end{itemize}

\subsection{Image Description Prompts}
\label{apdx:image_desc}
For images without native alt-text, we generate detailed descriptions using a high-resource pivot language to leverage stronger visual reasoning. We design domain-aware prompt templates (e.g., natural images, diagrams) and iteratively refine them via human inspection. Specifically, we randomly sample 200 images and ask reviewers to check a set of quality aspects (e.g., completeness, object relations, and factual alignment). If issues are found, we revise the prompt and regenerate until the sampled outputs meet the quality criterion.

Figure~\ref{fig: image-desc-prompt} shows an example prompt used to elicit detailed, structured descriptions in English.

\begin{figure}[!ht]
\begin{minipage}{\linewidth}
\fbox{ % 创建带黑色边框的文本框
\parbox[c][20cm][c]{\dimexpr\linewidth-5\fboxsep-5\fboxrule}{ % 设置文本框的内边距和高度
\small % 将文字字体大小设置为小一号
Please carefully observe the image from a specific lesser-known language country. Based on the main elements and scene in the image, generate a detailed and precise English description. Your description should focus on the following aspects:

1. Clearly describe the main subject of the image, such as people, specific objects, or key locations/scenes.

2. Describe the activity the subject is engaged in, along with its characteristics and condition, as well as how the subject is presented in a specific time and space.

3. Describe other elements in the image, and the spatial relationships and interactions between them and the main subject.

4. Incorporate background knowledge to elaborate on relevant cultural features of the country using the lesser-known language, such as traditional clothing, language, festivals, or customs.

5. Discuss how this culture is represented and symbolized in the image, and deepen understanding by connecting with text information.

6. You may extend your expression by drawing on personal experience, knowledge, or associations, covering artistic style, aesthetic preferences, cultural perceptions, etc.,  while respecting the culture itself.

7. Express your understanding of how the culture is visually conveyed in the image, highlighting its distinctive characteristics, enhancing visual impression, and helping the reader form a clear and vivid perception.

\#\# General

- What details in the image draw attention? What might these details signify?  

- Based on the context of the lesser-known language country, what emotions or messages does the image convey?

- If the action or situation in the image were to continue, what might happen next?  

\#\# Comprehensive

- What are the main objects or scenes shown in the image? 

- What role do these elements play? How are they positioned or spatially related to each other?  

\#\# People

- Who are the people in the image? What might their relationships be? What are they doing?  

- Consider their clothing, facial expressions, gestures, and postures. What do these convey?  

\#\# Culture

- What cultural elements are reflected in the image?  

- What aesthetic or symbolic meanings might these cultural elements carry?  

- How does the image reflect the unique traditional or modern aspects of the local culture of the lesser-known language country?  

\#\# History

- Does the image reference any historical events, symbols, or figures?  

- How is the image related to the historical development of the lesser-known language country?
}
}
\end{minipage}
\caption{An example of image description prompt. For different categories, we emphasize different aspects in the prompts.}
\label{fig: image-desc-prompt}
\end{figure}

\subsection{Quality Assurance and Filtering}
\label{apdx:qa}

To ensure the robustness and reliability of \datasetname, we apply a multi-stage quality assurance (QA) pipeline that combines automated filtering with human verification.

\paragraph{Automated Language Filtering.}
We apply strict language identification using \textit{fastText}~\citep{joulin2016bag}. For each image--text pair, only samples whose detected language matches the target language with confidence greater than 0.8 are retained. This step removes residual English or cross-lingual noise commonly introduced during large-scale web crawling.

\paragraph{Image--Text Relevance Filtering.}
We discard samples with weak or degenerate image--text alignment, including captions consisting solely of URLs, navigation boilerplate, file names, or repeated template text. We additionally remove samples exhibiting encoding corruption or empty textual content.

\paragraph{Human Verification Protocol for Data Quality}
\label{apdx:hv}
% To further validate data quality, we adopt a human-in-the-loop verification strategy. For each language, native speakers review random subsets of the processed data (200 samples per batch). Annotators flag samples that fall into any of the following categories:
% (i) garbled or corrupted text,
% (ii) non-informative boilerplate captions,
% (iii) clear semantic mismatch between image and text.

% If the rejection rate in a batch exceeds 5\%, filtering rules are refined and the entire sub-dataset is re-processed. This iterative procedure continues until the rejection rate consistently falls below the threshold, ensuring high-quality and culturally reliable supervision.

To ensure the quality of \datasetname, we employ a systematic human verification process in addition to automated filtering.

\textbf{Sampling strategy.}
For each target language, we randomly sample 200 image--text pairs from the processed dataset for manual inspection. 
Based on empirical experience with large-scale web-crawled datasets, this sample size is sufficient to reliably expose systematic quality issues, as problematic samples - when present - typically constitute a non-trivial portion of the data and are readily detectable within a 200-sample audit.

\textbf{Verification criteria.}
Native speakers of each target language manually review the sampled instances according to the following criteria:
\begin{itemize}
    \item Absence of meaningless, garbled, or corrupted text;
    \item No captions consisting solely of URLs or navigational boilerplate;
    \item Correct language identification;
    \item Semantic relevance and alignment between the image and the text;
    \item No apparent copyright or attribution violations.
\end{itemize}

\textbf{Iterative refinement.}
When quality issues are identified, the corresponding filtering rules or extraction procedures are refined and applied to the full dataset. A new random sample is then drawn for re-verification.

\textbf{Convergence criterion.}
This iterative verification process continues until no issues are observed in the sampled data, at which point the dataset is considered to meet our quality standards.

% =========================
% Appendix A: Evaluation details
% =========================
\section{Evaluation Details}
\label{apdx: evaluation details}

\subsection{Evaluation Prompts}
\label{apdx: Evaluation Prompts}
We use uniform prompts to ensure fair comparison across methods and languages. For samples from $D_{know}^{L}$, we use:
\begin{quote}
\textit{Describe the picture, and point out the people and objects in it.}
\end{quote}
For samples from $D_{ling}^{L}$, we use:
\begin{quote}
\textit{Describe this image.}
\end{quote}
For each target language $L$, we translate the English prompt into language $L$ using Google Translate, and keep prompts fixed across models.

\subsection{Automatic Metrics}
\label{apdx: Automatic Metrics}
We report standard captioning metrics on $D_{ling}$ as \emph{linguistic fluency sanity checks}, and use Keyword Accuracy on $D_{know}$ to quantify culture-grounded entity identification.

\paragraph{BLEU, METEOR, ROUGE-L.}
For $D_{ling}$, we compute BLEU~\citep{papineni-etal-2002-bleu}, METEOR, and ROUGE-L between the model output and the reference description. These metrics are used to check whether the model preserves well-formed generation in the target language under our instruction-following description style.

\paragraph{Keyword Accuracy on $D_{know}$.}
Keyword Accuracy measures whether the model output recovers salient culture-grounded entities that appear in native alt-text. For each language $L$, we extract a set of keywords from the reference alt-text using TF--IDF (after language-specific tokenization; Appendix~\ref{apdx:tokenization}) and keep the top-$K$ keywords per sample (we use a fixed $K$ across languages). A prediction is counted as correct if at least one reference keyword appears in the model output (exact match after normalization). We report the average accuracy over the test set.

\paragraph{Normalization and tokenization.}
We normalize text by lowercasing where applicable, removing punctuation, and applying language-specific tokenization (Appendix~\ref{apdx:tokenization}). This reduces spurious mismatches due to formatting.

\subsection{Tokenization and Language Processing}
\label{apdx:tokenization}
When tokenization is required for keyword extraction, we use \textit{pyarabic} for Arabic (AR), \textit{pythainlp} for Thai (TH), and \textit{stanza} for the remaining languages.

\subsection{Limitations of Keyword Accuracy}
\label{apdx: Limitations of Keyword Accuracy}
Keyword Accuracy is a language-agnostic proxy for culture-specific entity identification. It may underestimate semantically correct paraphrases or synonyms not captured by the extracted keyword list. We therefore complement automatic metrics with human evaluation (Appendix~\ref{apdx: human}) and qualitative case studies (Appendix~\ref{apdx:case_study}).

\begin{table}[!htbp]
\centering
\resizebox{0.5\columnwidth}{!}{%
\begin{tabular}{@{}cc@{}}
\toprule
 & average length \\ \midrule
mblip-mt0-xl & 91.26 \\
annotation & 1221.02 \\ \bottomrule
\end{tabular}%
}
\caption{Length of mblip-mt0-xl generation and MELLA annotation.}
\label{tab:mblip length}
\end{table}
% =========================
% Appendix: Experiment Details
% =========================
\section{Experiment Details}
\label{apdx: training details}

\subsection{Training Objective}
\label{apdx: training obj}
We adopt supervised fine-tuning (SFT) following prior work on multilingual and low-resource MLLM adaptation~\citep{zhang-etal-2024-enhancing-multilingual}. For each target language $L$, we fine-tune an existing MLLM in a parameter-efficient manner using the collected dataset $D^{L}=D^{L}_{ling}\cup D^{L}_{know}$.

To mitigate overfitting and encourage linguistic diversity, we manually craft a prompt pool of 20 prompts for each language:
\begin{equation}
P^{L} = \{x_i^{L} \mid i = 1, \ldots, 20\}.
\end{equation}
For each instance, a prompt $x \in P^{L}$ is randomly sampled. Given input image $I$ and prompt $x$, the model is trained to generate target text $T$ in language $L$ by minimizing standard cross-entropy:
\begin{equation}
\mathcal{L}_{\text{CE}} = -\mathbb{E}_{((I, x), T) \sim D^{L}}
\left[ \sum_{t=1}^{|T|} \log P_\theta(T_t \mid T_{<t}, I, x) \right],
\end{equation}
where $P_\theta$ denotes the model parameterized by $\theta$.

\begin{table*}[t]
\centering
\resizebox{\textwidth}{!}{%
\begin{tabular}{@{}llll|lll@{}}
\toprule
\textbf{Model} & \textbf{Setting} & \textbf{Hyperparameter} & \textbf{Value} & \textbf{Setting} & \textbf{Hyperparameter} & \textbf{Value} \\ \midrule
\multirow{8}{*}{InternVL2-8B} & \multirow{8}{*}{Main} & Epoch & 1 & \multirow{8}{*}{Ablation} & Epoch & 1 \\
 &  & Batch Size per GPU & 2, 4 &  & Batch Size per GPU & 4 \\
 &  & Learning Rate & 4e-5 &  & Learning Rate & 4e-5 \\
 &  & Warmup Ratio & 0.03 &  & Warmup Ratio & 0.03 \\
 &  & LR Scheduler Type & cosine &  & LR Scheduler Type & cosine \\
 &  & Weight Decay & 0.01 &  & Weight Decay & 0.01 \\
 &  & Max Seq Length & 4096 &  & Max Seq Length & 4096 \\
 &  & Gradient Accumulation Steps & 2 &  & Gradient Accumulation Steps & 2 \\ \midrule
\multirow{8}{*}{Qwen2-VL-7B-Instruct} & \multirow{8}{*}{Main} & Epoch & 1 & \multirow{8}{*}{Ablation} & Epoch & 1 \\
 &  & Batch Size per GPU & 4 &  & Batch Size per GPU & 4 \\
 &  & Learning Rate & 4e-5 &  & Learning Rate & 4e-5 \\
 &  & Warmup Ratio & default &  & Warmup Ratio & default \\
 &  & LR Scheduler Type & default &  & LR Scheduler Type & default \\
 &  & Weight Decay & default &  & Weight Decay & default \\
 &  & Max Seq Length & 4096 &  & Max Seq Length & 4096 \\
 &  & Gradient Accumulation Steps & default &  & Gradient Accumulation Steps & default \\ \bottomrule
\end{tabular}%
}
\caption{Hyperparameters used for InternVL2-8B and Qwen2-VL-7B-Instruct. ``default'' follows HuggingFace Trainer defaults.}
\label{tab:hyperparam}
\end{table*}

\subsection{Hyperparameters}
\label{sec:hyper}
Table~\ref{tab:hyperparam} lists the main hyperparameters used in the fine-tuning process. For training Qwen2-VL-7B-Instruct, we use Huggingface Trainer. 

\subsection{Prompt Pool}
\label{apdx:prompt_pool}
\begin{figure}[!ht]
    \centering
    \includegraphics[width=0.9\linewidth]{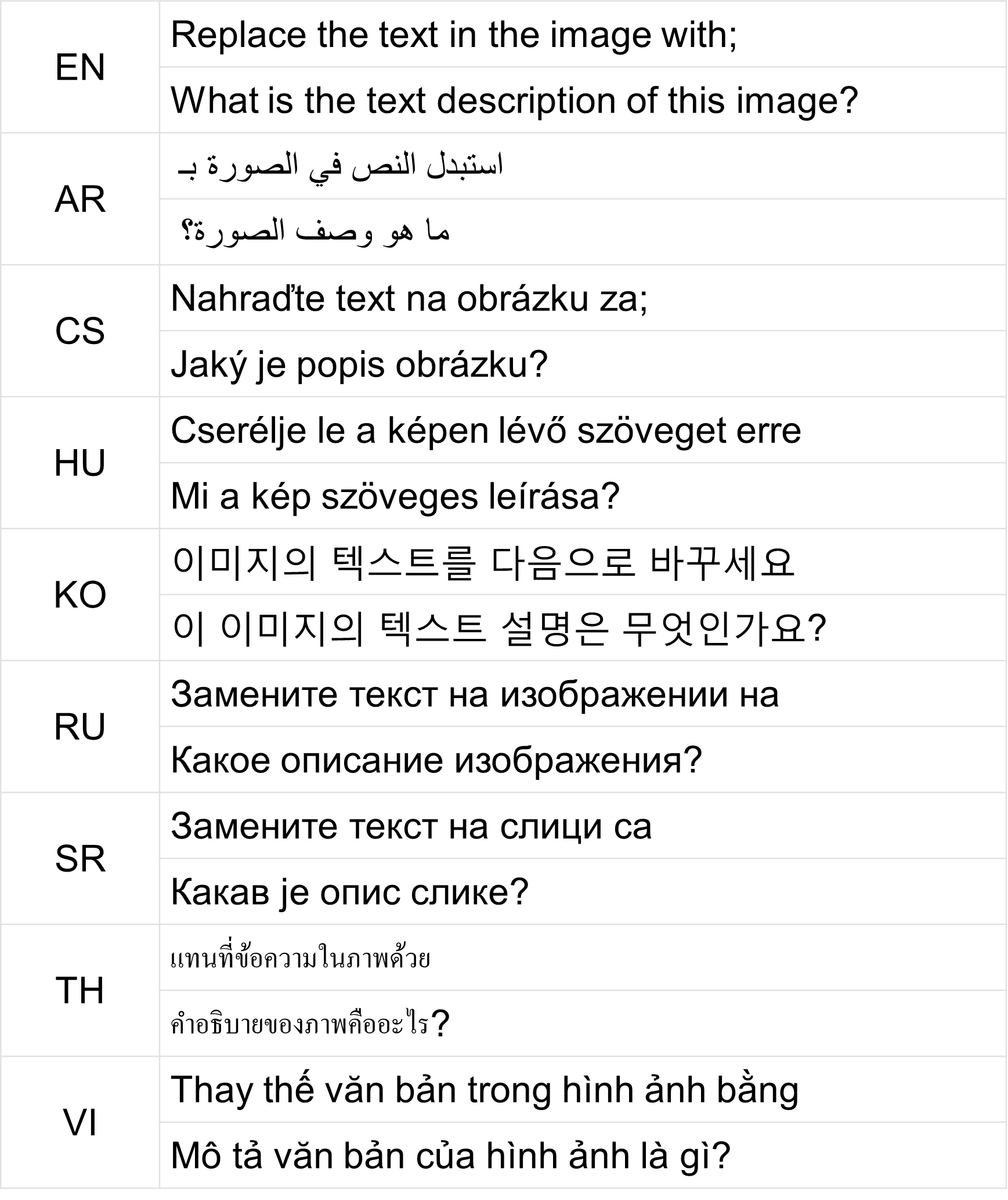}
    \caption{A subset of the prompt pool used for training.}
    \label{fig:prompt_pool}
\end{figure}

Figure~\ref{fig:prompt_pool} shows a subset of the manually designed prompt pool.

\subsection{Training Data Statistics}
\label{apdx:training_data_statistics}
Table~\ref{tab:training_data} reports the training sample counts used for each language and source.
\begin{table*}[t]
\centering
\small
\resizebox{\textwidth}{!}{%
\begin{tabular}{l|cccccccc}
\toprule
& AR & SR & RU & CS & KO & TH & VI & HU \\
\midrule
\textbf{InternVL2-8B} $D_{know}$ & 80K & 84.32K & 81K & 90K & 90K & 90K & 90K & 90K \\
\textbf{InternVL2-8B} $D_{ling}$ & 36K & 40.5K & 17.92K & 40.5K & 34.74K & 22.5K & 38.25K & 37.26K \\
\textbf{Qwen2-VL-7B} $D_{know}$ & 80K & 105.4K & 63K & 90K & 90K & 90K & 90K & 90K \\
\textbf{Qwen2-VL-7B} $D_{ling}$ & 36K & 40.5K & 26.25K & 40.5K & 34.74K & 22.5K & 38.25K & 37.26K \\
\bottomrule
\end{tabular}
}
\caption{Statistics of $D_{know}$ and $D_{ling}$ used in training.}
\label{tab:training_data}
\end{table*}

% =========================
% Appendix B: XM3600 sanity check
% =========================
\section{External Benchmark Sanity Check: XM3600}
\label{apdx:xm3600}
We additionally evaluate on XM3600~\citep{thapliyal2022crossmodal3600massivelymultilingualmultimodal} as an external sanity check. XM3600 consists of short, generic captions (typically 10--15 words), whereas \datasetname is designed for longer, instruction-following descriptions (typically 250+ words). This mismatch in target style can penalize models optimized for richer descriptions when using n-gram-based metrics.

Despite this difference in caption style, the results in Table~\ref{tab:xm3600_res} indicate that fine-tuning on \datasetname generally maintains comparable performance and often transfers positively on several languages, suggesting that our culture-grounded supervision does not harm general multilingual captioning ability.

\begin{table*}[h]
\centering
\small
\resizebox{0.8\textwidth}{!}{
\begin{tabular}{llccccccc}
\toprule
\textbf{Metric} & \textbf{Model} & \textbf{AR} & \textbf{TH} & \textbf{KO} & \textbf{CS} & \textbf{HU} & \textbf{RU} & \textbf{VI} \\
\midrule
\multirow{2}{*}{METEOR} & InternVL2-8B & 13.41 & 12.80 & 5.70 & 1.54 & 1.25 & 4.62 & 14.18 \\
 & + \datasetname & 13.91 & 13.90 & 7.38 & 2.08 & 3.70 & 3.44 & 11.34 \\
\midrule
\multirow{2}{*}{BLEU} & InternVL2-8B & 0.27 & 0.60 & 0.12 & 0.31 & 0.33 & 0.73 & 1.63 \\
 & + \datasetname & 0.54 & 0.04 & 0.38 & 0.54 & 0.57 & 1.02 & 1.50 \\
\midrule
\multirow{2}{*}{ROUGE-L} & InternVL2-8B & 1.76 & 0.30 & 1.03 & 2.20 & 2.03 & 4.86 & 14.18 \\
 & + \datasetname & 2.81 & 0.23 & 2.20 & 3.07 & 3.55 & 5.45 & 6.00 \\
\bottomrule
\end{tabular}
}
\caption{Evaluation on XM3600 as an external sanity check.}
\label{tab:xm3600_res}
\end{table*}

\section{Experiment results}
\label{Experiment results}

\subsection{Full main results}
\label{apdx:full res}
Table~\ref{tab: main results2} shows the full table of results. 
\begin{table*}[t]
\small
\centering
\resizebox{0.95\textwidth}{!}{%
\begin{tabular}{cccccccccc}
\toprule
\rowcolor[HTML]{F2F3F5} 
\textbf{Backbones} & \textbf{} & \textbf{AR} & \textbf{SR} & \textbf{RU} & \textbf{CS} & \textbf{KO} & \textbf{TH} & \textbf{VI} & \textbf{HU} \\ \midrule
% \rowcolor[HTML]{F0F4FF} 
% \multicolumn{10}{c}{\cellcolor[HTML]{F0F4FF}\textit{\textbf{Keyword Accuracy}}} \\
%  & - & 2.46 & 0.56 & 1.24 & 1.10 & 0.50 & 3.72 & 0.78 & 4.39 \\
%  & SDRRL & 2.39 & 0.33 & 1.22 & 1.37 & 1.02 & 3.38 & 1.00 & 2.00 \\
% \multirow{-3}{*}{InternVL2-8B} & \datasetname & \textbf{6.26} & \textbf{3.07} & \textbf{8.37} & \textbf{15.56} & \textbf{5.06} & \textbf{4.50} & \textbf{2.50} & \textbf{5.57} \\ \cmidrule(l){2-10} 
 % & - & 4.54 & 0.63 & 1.06 & 2.79 & 1.23 & 8.07 & 1.72 & 4.09 \\
 % & SDRRL &  &  &  &  &  &  &  &  \\
% \multirow{-3}{*}{InternVL2-26B} & \datasetname & \textbf{7.61} & \textbf{} & \textbf{} & \textbf{19.22} & \textbf{10.36} & \textbf{} & \textbf{} & \textbf{6.89} \\ \cmidrule(l){2-10} 
%  & - & 1.56 & 0.80 & 3.12 & 2.89 & 2.00 & 4.55 & 0.32 & 2.16 \\
%  & SDRRL & 0.01 & 0.66 & 0.45 & 1.78 & 0.01 & 2.86 & 0.15 & 1.57 \\
% \multirow{-3}{*}{Qwen2-VL-7B-Instruct} & \datasetname & \textbf{2.23} & \textbf{1.13} & \textbf{3.26} & \textbf{4.90} & \textbf{4.13} & \textbf{4.97} & \textbf{0.65} & \textbf{2.92} \\ \midrule
\rowcolor[HTML]{F0F4FF} 
\multicolumn{10}{c}{\cellcolor[HTML]{F0F4FF}\textit{\textbf{Meteor}}} \\
 & - & 26.07 & 2.70 & \textbf{7.71} & 3.37±0.66 & 14.54 & 19.95 & \textbf{18.19} & 0.11 \\
 & SDRRL & 22.46 & 5.23 & 5.83 & 6.62 & 13.83 & 11.77 & 11.1 & 5.68 \\
\multirow{-3}{*}{InternVL2-8B} & \datasetname & \textbf{29.78±0.49} & \textbf{13.54±0.42} & 4.91±0.50 & \textbf{12.17} & \textbf{22.81±0.31} & \textbf{22.5±0.72} & 16.37±0.55 & \textbf{13.11±0.31} \\ \cmidrule(l){2-10} 
 % & - & 20.42 & 3.87 &  & 5.95 & 16.56 &  &  & 5.39 \\
 % & SDRRL &  &  &  &  &  &  &  &  \\ 
% \multirow{-3}{*}{InternVL2-26B} & \datasetname & \textbf{14.97} & \textbf{} & \textbf{} & \textbf{10.05} & \textbf{19.28} & \textbf{} & \textbf{} & \textbf{11.55} \\
% \cmidrule(l){2-10} 
 & - & 15.49 & 2.33 & \textbf{6.54} & 6.03 & 12.93 & 17.14 & 16.77 & 6.37 \\
 & SDRRL & 2.35 & 0.25 & 1.28 & 5.32 & 0.76 & 18.48 & 1.92 & 7.01 \\
\multirow{-3}{*}{Qwen2-VL-7B-Instruct} & \datasetname & \textbf{36.89±0.35} & \textbf{13.88±0.57} & 5.36±0.77 & \textbf{12.88±0.40} & \textbf{23.74±0.55} & \textbf{34.63±0.67} & \textbf{28.66±0.48} & \textbf{12.72±0.43} \\ \midrule
% \rowcolor[HTML]{F0F4FF} 
% \multicolumn{10}{c}{\cellcolor[HTML]{F0F4FF}\textit{\textbf{BLEU}}} \\
%  & - & 1.79 & 1.05 & 5.56 & 1.31 & 2.56 & 0.15 & 6.91 & 0.05 \\
%  & SDRRL & 12.18 & 6.11 & \textbf{7.01} & 7.59 & 6.91 & 0.45 & 11.07 & 6.09 \\
% \multirow{-3}{*}{InternVL2-8B} & \datasetname & \textbf{13.96} & \textbf{13.22} & 4.40 & \textbf{14.33} & \textbf{11.02} & \textbf{0.56} & \textbf{15.53} & \textbf{13.45} \\ \cmidrule(l){2-10} 
% \cmidrule(l){2-10} 
%  & - & 2.45 & 0.60 & 3.24 & 2.37 & 1.48 & 0.32 & 8.17 & 3.40 \\
%  & SDRRL & 1.43 & 0.21 & 6.16 & 6.29 & 0.49 & 0.67 & 1.66 & 7.44 \\
% \multirow{-3}{*}{Qwen2-VL-7B-Instruct} & \datasetname & \textbf{19.95} & \textbf{16.33} & \textbf{6.26} & \textbf{14.80} & \textbf{11.48} & \textbf{1.00} & \textbf{30.18} & \textbf{13.39} \\ \midrule 
\rowcolor[HTML]{F0F4FF} 
\multicolumn{10}{c}{\cellcolor[HTML]{F0F4FF}\textit{\textbf{Rouge-L}}} \\
 & - & 5.23 & 6.41 & \textbf{12.73} & 6.25 & 6.25 & 0.50 & 12.39 & 0.22 \\
 & SDRRL & 14.37 & 7.07 & 8.60 & 10.18 & 9.17 & 1.55 & 9.98 & 7.91 \\
\multirow{-3}{*}{InternVL2-8B} & \datasetname & \textbf{17.26±0.33} & \textbf{18.77±0.39} & 6.32±0.81 & \textbf{17.74±0.60} & \textbf{14.97±0.71} & \textbf{2.25±0.32} & \textbf{14.57±0.72} & \textbf{18.41±0.50} \\ \cmidrule(l){2-10} 
  % & - & 8.63 & 7.87 &  & 10.54 & 8.54 &  &  & 9.71 \\
 % & SDRRL &  &  &  &  &  &  &  &  \\ 
% \multirow{-3}{*}{InternVL2-26B} & \datasetname & \textbf{9.19} & \textbf{} & \textbf{} & \textbf{14.76} & \textbf{12.90} & \textbf{} & \textbf{} & \textbf{16.45} \\
 \cmidrule(l){2-10} 
 & - & 11.30 & 5.50 & \textbf{12.86} & 1.11 & 7.85 & 1.31 & 16.84 & 11.30 \\
 & SDRRL & 1.59 & 0.38 & 10.19 & 8.38 & 0.87 & 2.22 & 1.82 & 10.29 \\
\multirow{-3}{*}{Qwen2-VL-7B-Instruct} & \datasetname & \textbf{24.13±0.30} & \textbf{20.08±0.57} & 8.47±0.63 & \textbf{19.02±0.58} & \textbf{16.08±0.89} & \textbf{3.31±0.38} & \textbf{27.45±0.61} & \textbf{18.51±0.45} \\
\bottomrule
\end{tabular}%
}
\caption{Main results of evaluating the understanding capabilities of MLLMs in the contexts of low-resource languages. Please note that ``\textit{Keyword Accuracy}'' is employed for evaluation on \(D_{know}\). ``\textit{BLEU}'',``\textit{Rouge-L}'' and  ``\textit{Meteor}'' are employed for evaluation on \(D_{ling}\).} 
\label{tab: main results2}
\end{table*}

% \subsection{Further evaluation}
% We evaluate our method on XM3600~\citep{thapliyal2022crossmodal3600massivelymultilingualmultimodal}, which is a multilingual image-caption dataset. 

\subsection{Comparison with mBLIP}
\label{apdx: mblip}
As mBLIP employs a similar machine translation-based approach to ours, we conduct a comparative evaluation against mblip-mt0-xl on our test sets. Table~\ref{tab:mblip} presents the results, revealing that mblip-mt0-xl exhibits limited performance on our evaluation benchmark. We attribute this performance gap to the characteristics of mBLIP's training data, which consists of considerably shorter captions with less detailed descriptions, as evidenced in Table~\ref{tab:mblip length}. This suggests that training data complexity and caption length play crucial roles in model performance on our tasks.

\begin{table*}[]
\resizebox{\textwidth}{!}{%
\begin{tabular}{@{}ccccccccc@{}}
\toprule
\rowcolor[HTML]{F2F3F5} 
\textbf{Backbones} & \textbf{RU} & \textbf{KO} & \textbf{TH} & \textbf{VI} & \textbf{AR} & \textbf{CS} & \textbf{SR} & \textbf{HU} \\ \midrule
\rowcolor[HTML]{F0F4FF} 
\multicolumn{8}{c}{\cellcolor[HTML]{F0F4FF}\textit{\textbf{METEOR}}} \\
mblip-mt0-xl & 0.96 & 1.73 & 1.81 & 7.61 & 10.53 & 0.75 & 1.49 & 1.54 \\
InternVl2-8B +MELLA & 4.91 (+3.95) & 22.81 (+21.08) & 22.50 (+20.69) & 16.37 (+8.76) & 29.78 (+19.25) & 12.17 (+11.42) & 13.54 (+12.05) & 13.11 (+11.57) \\
\multicolumn{1}{l}{Qwen2-VL-7B-Instruct +MELLA} & \multicolumn{1}{l}{5.36 (+4.40)} & \multicolumn{1}{l}{23.74 (+22.01)} & 34.63 (+32.82) & \multicolumn{1}{l}{28.66 (+21.05)} & \multicolumn{1}{l}{36.89 (+26.36)} & \multicolumn{1}{l}{12.88 (+12.13)} & \multicolumn{1}{l}{13.88 (+12.39)} & \multicolumn{1}{l}{12.72 (+11.18)} \\ \midrule
\rowcolor[HTML]{F0F4FF} 
\multicolumn{8}{c}{\cellcolor[HTML]{F0F4FF}\textit{\textbf{Rouge-L}}} \\
mblip-mt0-xl & 2.92 & 1.85 & 0.10 & 9.89 & 8.08 & 2.23 & 4.13 & 3.61 \\
InternVl2-8B +MELLA & 6.32 (+3.40) & 14.97 (+13.12) & 2.25 (+0.44)& 14.57 (+4.68) & 17.26 (+9.18) & 17.74 (+15.51) & 18.77 (+14.64) & 18.41 (+14.80) \\
\multicolumn{1}{l}{Qwen2-VL-7B-Instruct +MELLA} & \multicolumn{1}{l}{8.47 (+5.55)} & \multicolumn{1}{l}{16.08 (+14.23)} & 3.31 (+1.50) & \multicolumn{1}{l}{27.45 (+17.56)} & \multicolumn{1}{l}{24.13 (+16.05)} & \multicolumn{1}{l}{19.02 (+16.79)} & \multicolumn{1}{l}{20.08 (+15.95)} & \multicolumn{1}{l}{18.51 (+14.90)} \\ \bottomrule
\end{tabular}%
}
\caption{Compare with mblip-mt0-xl.}
\label{tab:mblip}
\end{table*}

\subsection{Human evaluation results}
\label{apdx: Human evaluation results}
\begin{figure}[!]
    \centering
    \includegraphics[width=\linewidth]{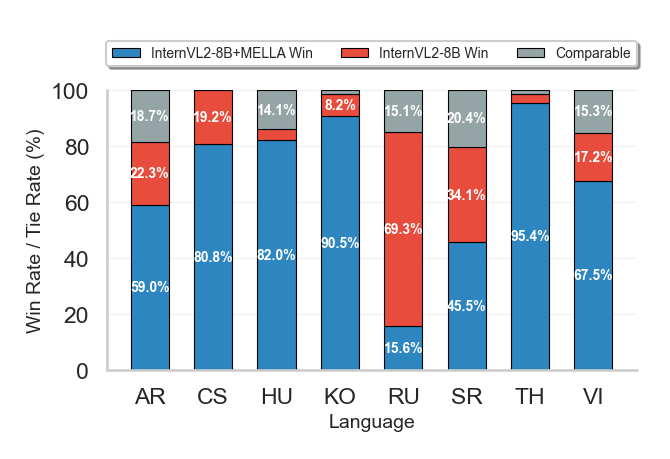}
    \caption{Human evaluation over 100 validation samples and 8 volunteers.}
    \label{fig: human eval}
\end{figure}
The overall human evaluation results, illustrated in Figure~\ref{fig: human eval}, show a strong alignment with our quantitative findings.

For the Russian (RU) subset in particular, human evaluators tended to prefer the pre-trained model over the fine-tuned one. This trend is consistent with the automatic evaluation results: as shown in Table~\ref{tab: main results2}, the BLEU score of InternVL2-8B slightly decreased after fine-tuning on RU data. Given that grammatical correctness was given the highest priority in our human evaluation criteria, this preference likely reflects the pre-trained model's stronger grammatical fluency, even though the fine-tuned model may have improved in content relevance or factual accuracy.

% =========================
% Appendix C: Cross-lingual transfer
% =========================
% \section{Cross-Lingual Transfer to High-Resource Languages}
% \label{apdx:cross_lingual}
% A common concern in multilingual fine-tuning is negative transfer or catastrophic forgetting in high-resource languages (e.g., English and Chinese). We therefore evaluate \datasetname-finetuned models on English (EN) and Chinese (ZH) benchmarks. Table~\ref{tab:cross_lingual} shows that fine-tuning on low-resource languages generally maintains (and occasionally improves) performance on EN/ZH relative to the baseline, suggesting limited negative transfer.

% \begin{table}[h]
% \centering
% \small
% \resizebox{0.9\linewidth}{!}{
% \begin{tabular}{lcccc}
% \toprule
% \textbf{Training Language} & \textbf{EN METEOR} & \textbf{EN BLEU-4} & \textbf{ZH METEOR} & \textbf{ZH BLEU-4} \\
% \midrule
% Baseline (No FT) & 17.37 & 2.01 & 16.23 & 0.00 \\
% \midrule
% FT on Arabic (AR) & 17.28 & 2.25 & 17.21 & 0.06 \\
% FT on Thai (TH) & 17.92 & 2.78 & 17.31 & 0.02 \\
% FT on Korean (KO) & 17.59 & 2.54 & 16.80 & 0.02 \\
% FT on Hungarian (HU) & 17.21 & 2.20 & 17.64 & 0.10 \\
% \bottomrule
% \end{tabular}
% }
% \caption{Cross-lingual transfer to EN/ZH. Fine-tuning on single low-resource languages maintains (and sometimes slightly improves) EN/ZH scores.}
% \label{tab:cross_lingual2}
% \end{table}

% =========================
% Appendix D: Performance variation / per-language notes
% =========================
\section{Per-Language Notes and Performance Variation}
\label{apdx:perf_variation}

\paragraph{Why performance varies across languages.}
We observe variations across languages that can be attributed to (i) typological differences and tokenization difficulty; (ii) differences in base-model pretraining coverage; and (iii) differences in the availability and quality of $D_{ling}$ and $D_{know}$ across languages. For example, Russian (RU) sometimes exhibits more modest changes than Czech (CS) or Hungarian (HU), plausibly because RU is better represented in pretraining, leaving less headroom and potentially increasing interference when culture-specific signals conflict with generalized priors.

\paragraph{Alt-text as knowledge-rich but linguistically sparse supervision.}
Training solely on $D_{know}$ can improve culture-specific entity identification but degrades linguistic fluency, consistent with the short, elliptical, and stylistically homogeneous nature of alt-text. Integrating $D_{know}$ with $D_{ling}$ balances knowledge grounding and linguistic richness, supporting the view that native alt-text is best leveraged as a complementary knowledge source rather than a standalone corpus.

% =========================
% Appendix E: More case studies + human eval protocol
% =========================
\section{Additional Qualitative Examples and Human Evaluation}
\label{apdx:qual_human}

\subsection{Human Evaluation Protocol}
\label{apdx: human}
To assess generation quality beyond automatic metrics, we conduct human evaluation on 100 sampled instances. Evaluators are blind to which system produced each output.

\paragraph{Evaluators.}
We recruit native speakers or individuals who lived in the corresponding language region for more than three years, as well as participants majoring in the respective language, to ensure linguistic proficiency and cultural familiarity.

\paragraph{Criteria.}
We apply a three-stage protocol:
\begin{enumerate}
    \item \textbf{Grammatical correctness}: if one output contains severe grammatical errors or garbled text, the other is preferred.
    \item \textbf{Content relevance}: if one output is irrelevant to the image, the other is preferred.
    \item \textbf{Content accuracy}: when both are fluent and relevant, the more accurate and culturally appropriate output is preferred.
\end{enumerate}
Evaluators may mark \emph{comparable} if both are of similar quality.

\subsection{Case Studies}
\label{apdx:case_study}

\begin{figure*}[t]
    \centering
    \begin{subfigure}[t]{0.8\textwidth}
        \centering
        \includegraphics[width=\textwidth]{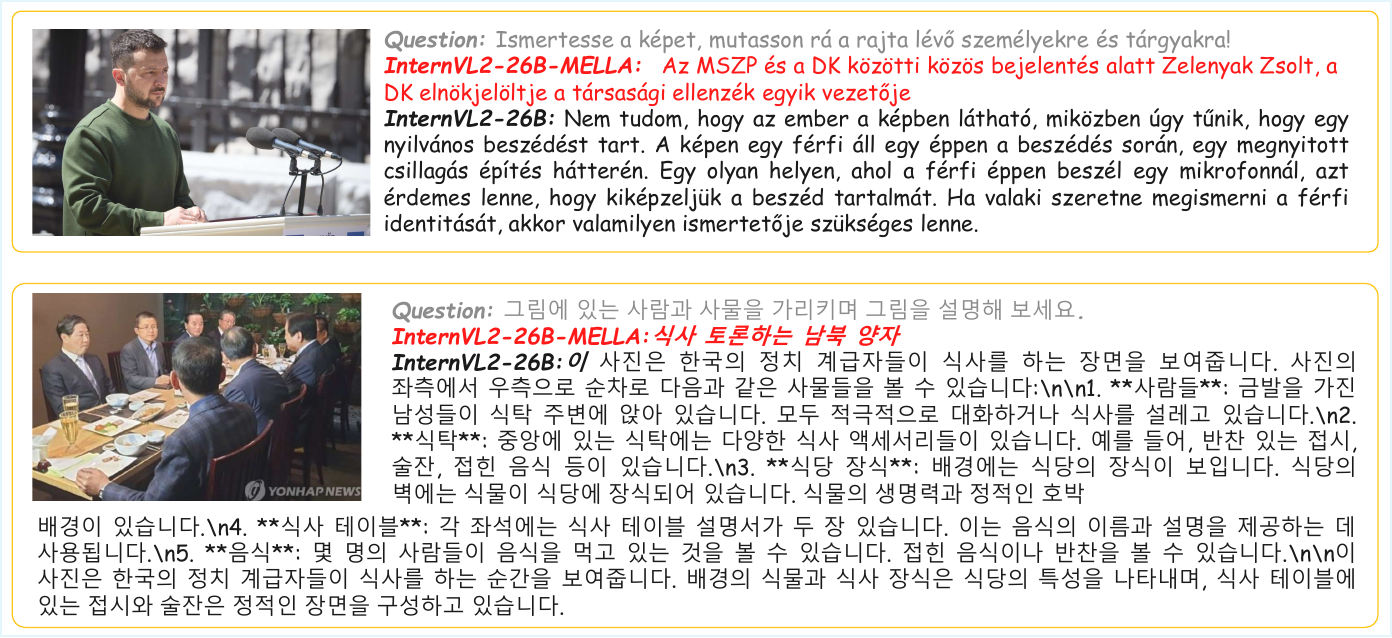}
        \caption{Cases on HU and KO. The cultural connotation has been recognized.}
        \label{fig:case1}
    \end{subfigure}
    \begin{subfigure}[t]{0.8\textwidth}
        \centering
        \includegraphics[width=\textwidth]{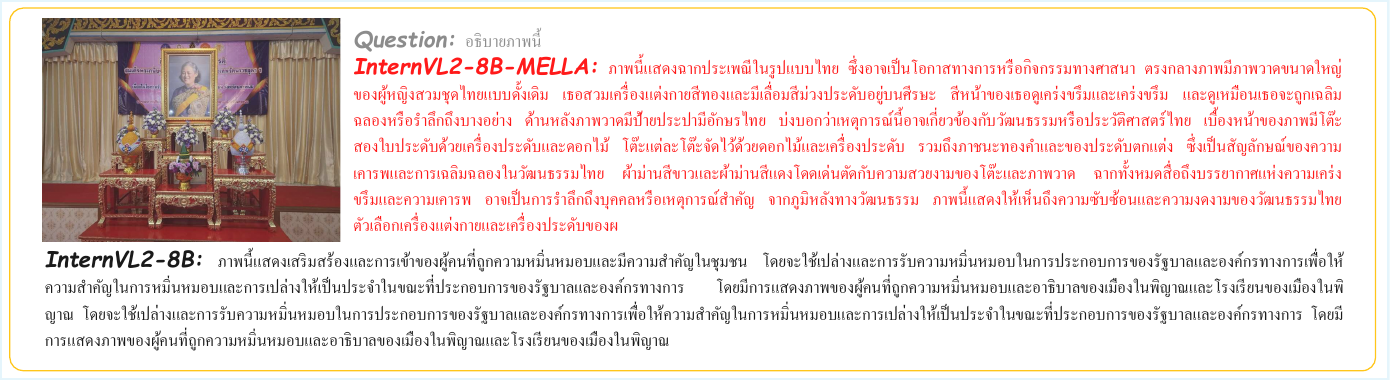}
        \caption{A case on TH. The issue of repeated outputs has been resolved, and Thai cultural elements have been incorporated into the descriptions.}
        \label{fig:case2}
    \end{subfigure}
    \begin{subfigure}[t]{0.8\textwidth}
        \centering
        \includegraphics[width=\textwidth]{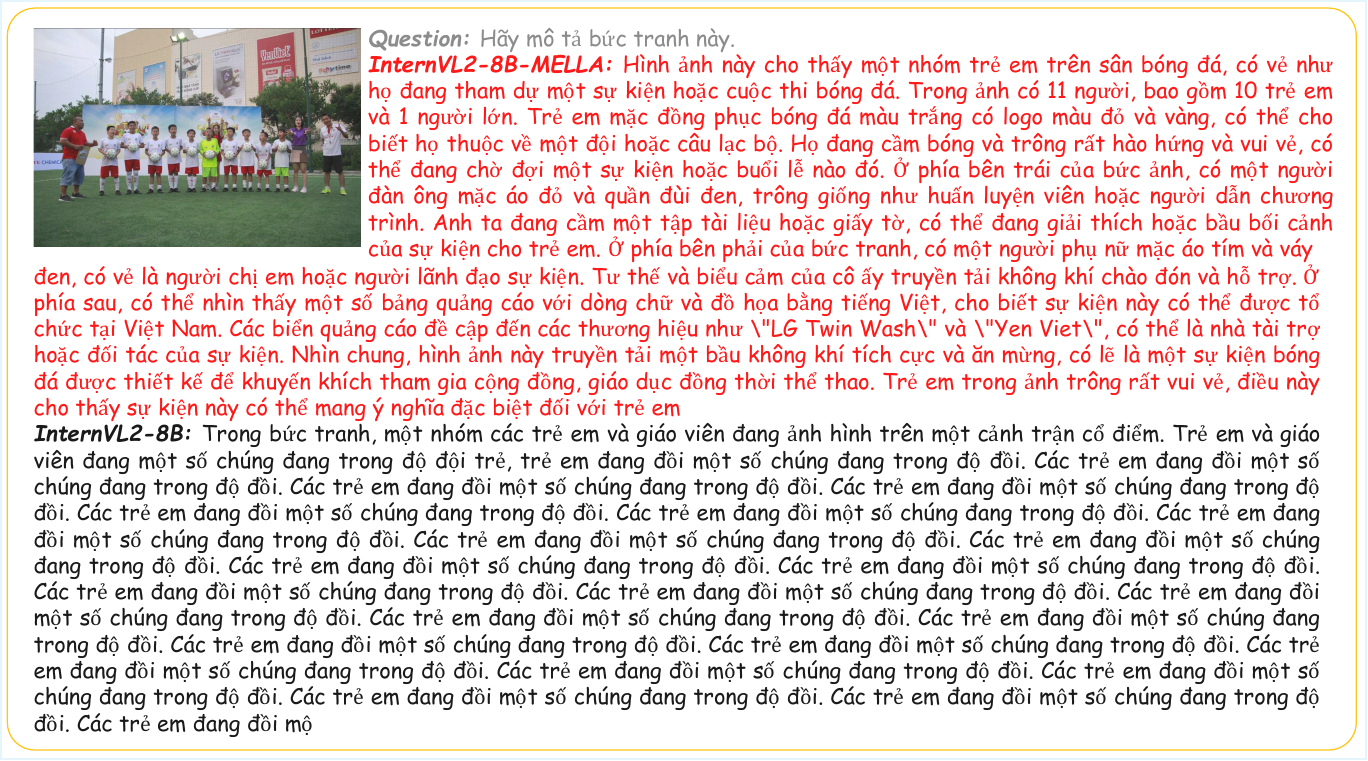}
        \caption{A case on VI. The issue of repeated outputs has been resolved.}
        \label{fig:case3}
    \end{subfigure}

    \caption{Case studies on HU, KO, TH, and VI showing improved cultural understanding and resolution of repeated output issues.}
    \label{fig:case_all}
\end{figure*}

Figure~\ref{fig:case_all} presents additional case studies. These examples clearly demonstrate how our training process enhances the MLLM's linguistic capabilities and cultural understanding. Although some hallucinations are observed - an inherent limitation of alt-text data~\citep{birhane2021multimodaldatasetsmisogynypornography} - our method serves as a strong example of the effectiveness of the proposed dual-source data strategy. Figure~\ref{fig:case1} shows our attempt at a larger model, which demonstrates a certain degree of cultural grounding. The examples in Figures~\ref{fig:case2} and~\ref{fig:case3} illustrate how the model incorporates culturally relevant knowledge when generating image descriptions.

\end{document}